Highlights

**BAR-Analytics: A Web-based Platform for Analyzing Information Spreading Barriers in News: Comparative Analysis Across Multiple Barriers and Events**

Abdul Sittar, Dunja Mladenić, Alenka Guček, Marko Grobelnik

- BAR-Analytics is an open-source, web-based platform for comprehensive news-event analysis using multiple analytical approaches.

- It collects barrier-related metadata (economic, political, geographical, cultural) for contextual analysis across diverse news sources and regions.

- It provides four core analytical capabilities: Propagation Analysis, Trend Analysis, Hierarchical Topic Modeling, and Sentiment Analysis.

- A demonstration of BAR-Analytics applied to the Russian-Ukrainian and Israeli-Palestinian conflicts reveals differences in news coverage across various barriers.

# BAR-Analytics: A Web-based Platform for Analyzing Information Spreading Barriers in News: Comparative Analysis Across Multiple Barriers and Events

Abdul **Sittar**[a,*,1], Dunja **Mladenić**[b,1], Alenka **Guček**[c,1] and Marko **Grobelnik**[d,1]

[a]*Jožef Stefan Institute, Jamova cesta 39, Ljubljana 1000, , Slovenia*

## ABSTRACT

This paper presents BAR-Analytics, a web-based, open-source platform designed to analyze news dissemination across geographical, economic, political, and cultural boundaries. Using the Russian-Ukrainian and Israeli-Palestinian conflicts as case studies, the platform integrates four analytical methods: propagation analysis, trend analysis, sentiment analysis, and temporal topic modeling. Over 350,000 articles were collected and analyzed, with a focus on economic disparities and geographical influences using metadata enrichment. We evaluate the case studies using coherence, sentiment polarity, topic frequency, and trend shifts as key metrics. Our results show distinct patterns in news coverage: the Israeli-Palestinian conflict tends to have more negative sentiment with a focus on human rights, while the Russia-Ukraine conflict is more positive, emphasizing election interference. These findings highlight the influence of political, economic, and regional factors in shaping media narratives across different conflicts.

## 1. Introduction

The global dissemination of information is largely driven by the news media, which plays a pivotal role in reporting events of worldwide significance. However, some news events are covered only in specific geographic areas, primarily due to barriers that impede the flow of information (Wilke et al., 2012). The term "barrier" refers to the conceptual divides that exist between different societies, nations, and regions during the transmission of information. Such obstacles manifest themselves in multiple forms, influencing the selection, reporting, and dissemination of news.

Several factors contribute to these barriers, including cultural context, political alignment, economic conditions, geographical proximity, and linguistic similarities. Domestic factors primarily shape news related to local occurrences, whereas national and international dynamics influence the coverage of global events. These factors - economic, political, cultural, linguistic and geographic - affect the flow of information differently depending on the nature of the event (Sittar et al., 2022).

News stories are often anchored to specific times, places, or entities, and the preferences of news publishers significantly affect coverage (Rospocher et al., 2016; Sittar et al., 2022). The interplay between political actors and journalists shapes news production, while the spread of fake news is often politically motivated Martens et al. (2018), (Zhang et al., 2025). Furthermore, economic disparities and political stability influence not only the selection of news but also its propagation and analysis (Chang and Lee, 1992).

Analyzing news barriers is crucial for understanding the challenges that affect the dissemination of information across different geographical, economic, political, and cultural contexts. Rather than simply classifying these barriers, it is essential to investigate how they shape news coverage, influence public perception, and impact the spread of information (Sittar and Mladenic, 2023). This analysis provides valuable insights for various real-world applications, such as: understanding news dissemination patterns and how information flows across different regions and societal structures, detecting misinformation and bias by identifying the underlying barriers that distort factual reporting, improving content recommendations by considering the influence of political, cultural, and economic barriers on audience preferences, optimizing content distribution by recognizing the factors that determine which news reaches specific demographics, and enhancing information policy and regulation to ensure a more balanced and transparent media landscape. For instance, consider a scenario where a news story circulates in a country with a strong cultural resistance to vaccinations. If an article claims a new vaccine is completely safe and effective, it may face skepticism, as cultural biases shape public perception. Readers might reject or question the credibility of the news, not necessarily because of its accuracy, but due to pre-existing societal barriers. By analyzing such barriers, we can better understand how misinformation spreads and the conditions that facilitate or hinder factual reporting.

An essential factor in this context is the sentiment of news regarding various events across different locations. Numerous studies have utilized sentiment analysis from textual sources, including social media and news articles, to predict financial variables (Consoli et al., 2022; Kumbure et al., 2022). The emotional tone of news significantly influences its propagation, as (Bustos et al., 2011) found a direct

---

*Corresponding author

abdul.sittar@ijs.si (A. Sittar); dunja.mladenic@ijs.si (D. Mladenić); alenka.gucek@ijs.si (A. Guček); marko.grobelnik@ijs.si (M. Grobelnik)

(A. Sittar); (D. Mladenić); (A. Guček); (M. Grobelnik)
ORCID(s):

correlation between stock market price movements and news dissemination patterns. Likewise, global events elicit varying sentiment polarities across geographical barriers, a concept (Moreo et al., 2012) refers to as the measurement of news popularity in a global context. Additionally, market behavior can be anticipated based on sentiment analysis Shah et al. (2018), which can differ depending on demographic factors, news sources, and geographical regions (Mehler et al., 2006).

Spatio-temporal analysis serves as a powerful tool for uncovering relationships between different locations over time (Sittar et al., 2021). For instance, (Ghosh and Cartone, 2020) explore spatial effects and spatio-temporal patterns associated with the COVID-19 outbreak across various regions in Italy. Similarly, (Paez et al., 2021) identify seasonal variations in disease patterns in Spain, attributing these fluctuations to changes in temperature, humidity, and sunlight exposure (Comito, 2021a). Moreover, (Gross et al., 2020) investigate the spatio-temporal spread of the initial wave of COVID-19 in China, comparing it with global trends based on factors like distance, population density, and human mobility. This analytical approach not only highlights how events evolve over time but also elucidates the interconnectedness of regions based on shared characteristics.

Understanding the differences in news reporting has gained increasing importance, particularly as the number of events with international implications continues to rise. Researchers and professionals in various fields, including digital humanities, media studies, and journalism, must examine these discrepancies. Recent significant events, such as the migration crisis in Europe, the COVID-19 pandemic, Brexit, the Russian-Ukrainian conflict, and the Israeli-Palestinian war, underscore the necessity of this analysis. Generally, newspapers tend to reflect the overall situation in specific areas, often neglecting broader contexts. The study of these differences enables a deeper comprehension of how political, economic, and cultural barriers shape public perception and influence the narrative around global events.

### 1.1. Contributions

The contributions of this research can be summarized as follows:

- We present a methodology for analyzing news dissemination across geographical, economic, political, and cultural boundaries, providing a comprehensive perspective on global news dynamics.

- We present BAR-Analytics, an open-source, web-based platform that fuses multiple analytical approaches for comprehensive news-event analysis.

- This platform offers four types of analyses: 1) Propagation Analysis: demonstrating how news spreads over time using a feed-forward mechanism, 2) Trend Analysis: comparing the frequency of news across various barriers, 3) Hierarchical Topic Modeling: illustrating differences in news reporting across different categories of barriers, and 4) Sentiment Analysis:

showing sentiment trends over time across various barrier categories.

- As a demonstration, we apply BAR-Analytics to the Russian-Ukrainian and Israeli-Palestinian conflicts. This use case highlights how the platform synthesizes various analyses, illustrating differences in news reporting across geographical, political, and economic divides while showcasing the platform's integrated functionality.

## 2. Related Work

Differences in news reporting across geographical, economic, political, and cultural barriers play a critical role in shaping public opinion and influencing global narratives. For example, Western media coverage of the Ukraine conflict frequently highlights humanitarian crises, civilian casualties, and geopolitical consequences, often portraying Russia as the aggressor and seeking international support for Ukraine Fisher (2023). In contrast, Russian media justify military intervention by framing Ukraine as unstable and emphasizing national security concerns Wang (2023). Similarly, in the Palestinian-Israeli conflict, Israeli media tend to focus on security threats and military responses, while international outlets emphasize civilian casualties and humanitarian issues Susser (2023). These contrasting narratives, shaped by barriers such as culture, geography, and politics, lead to polarized public perceptions, affecting how audiences interpret and respond to conflicts.

News dissemination barriers also significantly influence policy decisions. Governments often rely on media narratives to gauge public sentiment and shape strategies. Historical cases, such as media coverage of the Vietnam War, show how news reporting affected political and military decisions in the United States Scott (2008). Similarly, biased reporting on contemporary conflicts like Ukraine or Palestine can lead to misinformed policies, exacerbating tensions or delaying conflict resolution Liu (2019). Economic-driven media outlets may also frame conflicts in terms of financial impacts, influencing decisions on sanctions, military aid, or diplomatic interventions Okello (2023).

Furthermore, the psychological impact of news reporting varies across barriers. Reports emphasizing violence and suffering may increase stress and anxiety among soldiers and civilians, while overly optimistic portrayals can lead to unrealistic expectations Fry (2016); Huang (2010). These psychological effects underscore the importance of balanced and culturally sensitive reporting to avoid misinformation and emotional distress Li (2023).

Topic modeling (TM) has been widely used to analyze discussions around COVID-19, extracting popular topics and tracking their evolution over time (Älgå et al., 2020; Amara et al., 2021; Boon-Itt and Skunkan, 2020). BERTopic is a modern technique for topic inference from text documents, leveraging transformer-based embeddings and clustering methods. Unlike traditional approaches like Latent Dirichlet

Allocation (LDA), BERTopic has shown superior performance in producing coherent topics and handling complex textual data (AlAgha, 2021). Social media platforms, especially Twitter, often use pooling methods based on hashtags and authors to group documents, facilitating more accurate topic modeling. Carmela's approach, combining peak detection and clustering, is an example of using time series to identify and cluster spatio-temporal topics Comito (2021b). However, unlike social media, news articles lack such clear metadata for pooling, making it difficult to group articles effectively. Some studies, such as (Ghasiya and Okamura, 2021), have used algorithms like Top2Vec to overcome this by clustering similar news articles based on their content.

Sentiment classification identifies positive, negative, and neutral news, aiding in trend prediction across various domains. Previous studies have incorporated sentiment analysis alongside other features such as entities and phrases for tasks like news classification and fake news detection (Demirsoz and Ozcan, 2017; Hui et al., 2017; Yazdani et al., 2017). DistilBERT has demonstrated higher accuracy than traditional methods like TF-IDF in sentiment classification (Dogra et al., 2021). Additionally, sentiment-based approaches have been applied in stock market forecasting and fake news detection, where negative sentiments tend to spread rapidly (Ajao et al., 2019; Bhutani et al., 2019; Li et al., 2017).

Empirical and theoretical studies have explored how news narratives differ across regions. Content analysis of global news has found that Western media frame the Ukraine war as a struggle for democracy, while Russian media focus on NATO expansionism and historical ties. Similarly, the Palestine conflict is framed as self-defense by Israeli media, whereas Middle Eastern outlets highlight Palestinian resistance and suffering Fengler et al. (2020); Hunter and Liu (2006). Sentiment analysis shows that Western media tend to portray Israel more favorably, while Middle Eastern sources are more critical of Israeli actions McTigue (2011). Discourse analysis reveals linguistic and rhetorical differences, with Western media emphasizing international law and sovereignty in Ukraine, while Russian media focus on protecting ethnic Russians. In the Palestine conflict, Western narratives highlight security and terrorism, whereas Middle Eastern sources focus on occupation and liberation Salama et al. (2023).

The framing theory suggests that media outlets construct narratives aligned with national interests and audience expectations, often reducing complex issues into binary conflicts. Western media typically depict Ukraine as a battle between democracy and autocracy, while Russian outlets present it as resistance against Western imperialism Khamis and Dogbatse (2024). The propaganda model further examines how media ownership, government influence, and economic interests shape news reporting. Western media, often influenced by corporate and political ties, may downplay Palestinian perspectives, whereas Arab media emphasize Palestinian suffering due to regional solidarityTasseron

(2023).

Recent studies have introduced new methodologies for analyzing global news dissemination networks. (Alipour et al., 2024) reconstructed news propagation by analyzing 140 million articles from 183 countries using the GDELT database. This study incorporated a temporal dimension, tracking how news spreads across nations and identifying "news superspreaders"—a small group of influential countries that dominate global media flows. Using a gravity model, they found that economic power and geographical proximity drive international news connectivity, reinforcing existing global media imbalances.

While (Alipour et al., 2024) focused on connectivity patterns and influential spreaders, our study introduces a novel dimension by explicitly examining how different barriers (geographical, economic, political, and cultural) shape news propagation and reporting. Unlike previous work, BAR-Analytics integrates multiple analytical techniques—including sentiment analysis, trend analysis, hierarchical topic modeling, and news propagation analysis—within a unified platform. By applying this framework to real-world conflicts like Russia-Ukraine and Israel-Palestine, our study provides a more nuanced perspective on how economic and political contexts influence media narratives, an area largely unexplored in previous research.

Other studies, such as MIDDAG, examine information propagation, but with a focus on social media dynamics. MIDDAG visualizes how COVID-19-related news spreads across Twitter and Reddit, analyzing user engagement, event tracking, and forecasting (Ma et al., 2024). In contrast, BAR-Analytics shifts the focus to macro-level analysis, examining how news spreads across major geopolitical regions and how barriers shape content emphasis. Unlike MIDDAG, which primarily explores user interactions, BAR-Analytics compares media narratives across global contexts, uncovering systematic differences in news reporting (Ma et al., 2024).

Similarly, Tanbih and BAR-Analytics share a goal of enhancing media literacy, but they differ in scope. Tanbih primarily assesses media bias, factuality, and propaganda at the article and source level, whereas BAR-Analytics introduces spatio-temporal analysis to explore how news narratives evolve over time across geographical, economic, and political barriers (Zhang et al., 2019). This cross-barrier perspective allows for a comparative analysis of major geopolitical conflicts, a dimension largely missing from previous research.

Finally, while Media Cloud tracks global news content across sources, it does not explicitly analyze the influence of multiple barriers on news propagation. BAR-Analytics advances this research by incorporating geographical, economic, political, and cultural barriers into news dissemination analysis. By applying this approach to Russia-Ukraine and Israel-Palestine, our study extends the understanding of how news narratives are shaped by these contextual factors, addressing a gap in existing literature (Roberts et al., 2021). Despite extensive empirical and theoretical studies, barriers in global news reporting remain underexplored. Prior research has analyzed news connectivity, media bias, and audience reception, but few studies systematically compare barriers across multiple conflicts. BAR-Analytics fills this gap by offering an integrated platform for analyzing how different barriers influence news narratives. Unlike existing approaches that focus on single aspects of news dissemination, our work introduces a multi-dimensional analysis, providing a more comprehensive perspective on global news dynamics.

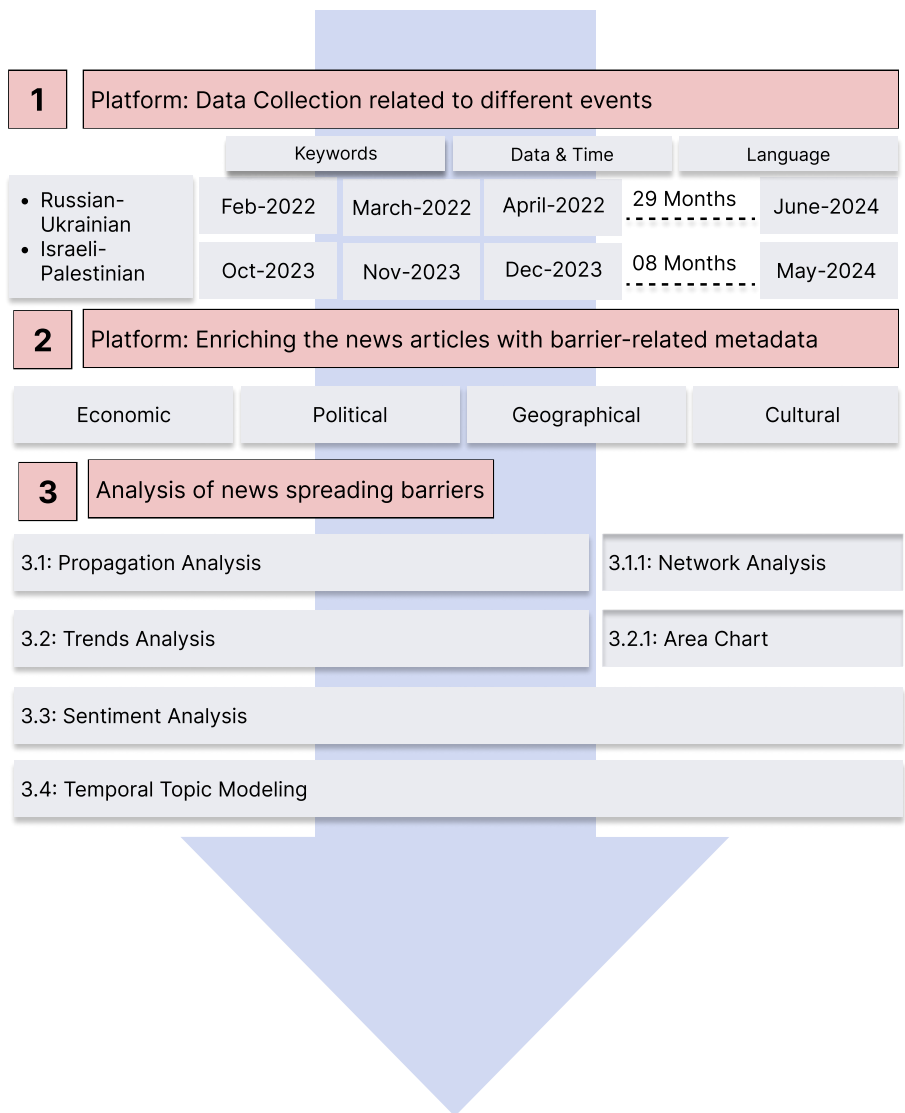

**Figure 1:** Overview of the methodology: (1) Data collection of war-related news articles, (2) Metadata enrichment with barriers information, and (3) Four-step analysis including propagation, trends, sentiment, and topic modeling.

## 3. Methodology

Our methodology outlines the process of collecting, enriching, and analyzing news articles related to two major conflicts: the Russian-Ukrainian war and the Israeli-Palestinian war. It details the steps of data collection, metadata enrichment, and four distinct analyses—information propagation, trend analysis, sentiment evaluation, and topic modeling—to examine how news spreads, its sentiment, and the influence of economic and geopolitical factors on news dissemination. This comprehensive approach provides valuable insights into the dynamics of news coverage during significant global events.

### 3.1. Data Collection

The first step involves data collection, where news articles were gathered for two major conflicts: the Russian-Ukrainian war and the Israeli-Palestinian war. Articles related to the Russian-Ukrainian conflict span from February 2022 to June 2024, while those covering the Israeli-Palestinian conflict range from October 2023 to May 2024. In total, 120,000 articles were collected for the Israeli-Palestinian war and 230,000 for the Russian-Ukrainian war, all in English. These articles were obtained through the Event Registry API s (Leban et al., 2014)[1], ensuring a comprehensive dataset that includes essential attributes such as title, body text, publisher name, publication date and time, DMOZ categories, and Wikipedia concepts. The selection process was guided by predefined Wikipedia categories relevant to each war, such as "Russo-Ukrainian War," "Russia-Ukraine Relations," "Russian Invasion of Ukraine," "Israel-Hamas War," "Gaza-Israel Conflict," and "Israeli-Palestinian Conflict". The peak in news articles in December 2023 likely reflects major developments in the Israeli-Palestinian war, drawing intense media attention. Over time, coverage declined as the initial urgency faded, media saturation set in, and public interest shifted. Additionally, unless new significant events occurred, reporting became more selective, leading to fewer published articles (see Figure 2). The surge in news articles in July 2023 related to the Russian-Ukrainian war likely corresponds to significant geopolitical events, military escalations, or diplomatic developments that attracted heightened media attention. The increasing trend in coverage over time suggests a sustained global interest, possibly driven by ongoing conflicts, international responses, or emerging crises. This pattern reflects how major events trigger spikes in reporting, followed by continued engagement as the situation evolves (see Figure 2).

---

[1] https://github.com/EventRegistry/event-registry-python/blob/master/eventregistry/examples/QueryArticlesExamples.py

## 3.2. Enriching the news articles with barrier-related metadata

The second step enriches the collected news articles with metadata related to barriers. A barriers database was created by compiling information from approximately 8,000 news publishers identified through the Event Registry. The primary metadata enrichment task involved determining the headquarters location of each news publisher, which was extracted from Wikipedia infoboxes using the Bright Data service. Since Wikipedia covers a substantial number of well-known publishers, this approach ensured a reliable metadata source. Additionally, the economic barrier was quantified using the Legatum Prosperity Index, which evaluates countries based on twelve economic dimensions. A numerical representation of economic disparity between countries was derived using the Euclidean distance function. To classify countries based on their economic characteristics, a k-means clustering algorithm was applied, resulting in 20 economic clusters. The resulting metadata, stored in the barriers database, facilitates an analysis of how different economic and geopolitical contexts influence news dissemination. A list of countries and their respective economic classifications is available in a GitHub repository and presented in an appendix (see Table 1).

## 3.3. Analysis of news spreading barriers

The final step comprises four distinct analyses applied to the collected and enriched news data. The first, information propagation analysis, examines how news about war events spreads over time. This is achieved by computing the cosine similarity between articles and subsequent publications to measure textual resemblance. The propagation patterns are visualized using a dynamic 3D network graph, where communities—clusters of highly related articles—are detected through the Girvan-Newman algorithm. Communities with a small number of articles typically contain near-identical reports published within a short timeframe, whereas larger communities may indicate broader discussions or evolving narratives.

The second analysis focuses on trends, capturing fluctuations in news coverage over time through line graphs. These visualizations highlight how different barriers influence the volume of reporting across various periods.

The third analysis evaluates sentiment within the news articles. By examining the textual content, this analysis determines whether news coverage conveys positive, negative, or neutral sentiment regarding war events. For sentiment analysis, a score is assigned to each article using VADER, a lexicon-based tool designed for social media sentiment analysis. VADER computes a sentiment score between -1 (negative) and 1 (positive), considering both word sentiment values and modifying rules such as intensifiers and negations. The score is derived from the first five sentences of an article, with values below -0.1 classified as negative, between -0.1 and 0.1 as neutral, and above 0.1 as positive (Leban et al., 2014; Taj et al., 2019).

The fourth analysis utilizes hierarchical topic modeling through BERTopic to identify key themes within the news corpus. This technique enables the categorization of dominant topics by leveraging transformer-based embeddings and clustering methods, offering insights into major issues across various barriers. The methodology involves several key steps: data preprocessing, where article text is cleaned using NLP techniques such as stopword removal and tokenization; topic extraction, where BERTopic fits and transforms the corpus to generate topic clusters based on topic frequency and word importance; and hierarchical clustering, where document similarity is computed using TF-IDF vectorization and cosine similarity, followed by Ward's linkage clustering to refine the topic structure. Topic labeling and visualization are then performed using BERTopic's topic generation function, producing visual representations like bar charts and dendrograms for better interpretability. The model's performance is evaluated using topic coherence metrics such as $C_{\text{npmi}}$, and $C_{\text{p}}$, ensuring meaningful and distinguishable topics. Additionally, topic diversity is verified by measuring unique terms across clusters, and qualitative validation is conducted through word cloud analysis, ensuring that the extracted topics provide a structured and interpretable representation of news discussions across different barriers.

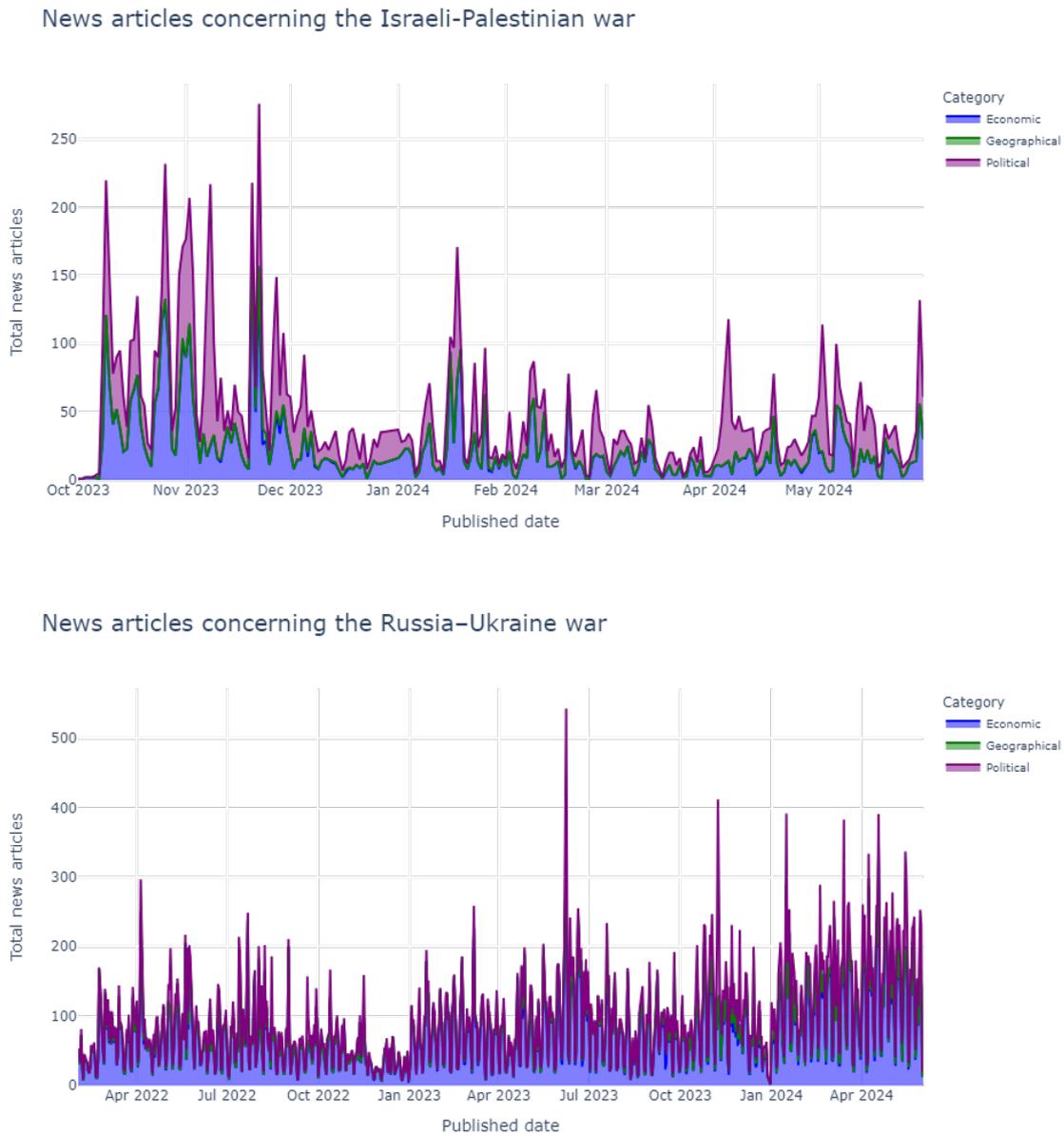

**Figure 2:** The line charts show the statistics about the news articles for the three barriers (Economic, Geographic, and Political) concerning the two wars Israeli-Palestinian war and Russia-Ukraine war respectively. Total news articles (y axis) are displayed for the period of Feb 2022-Jun 2024 for the Russian-Ukrainian war and Oct 2023-May 2024 for the Israeli- Palestinian war.

## 4. Proposed Framework

The BAR-Analytics platform is build based on the novel framework structured around three key components: 1) the creation of a barriers database, 2) the development of analytical methods to discern differences among barriers, and 3) the integration of these components into a web-based application.

### 4.1. Creating the Barriers Database

To develop the barriers database, we gathered the names of prominent news publishers from Event Registry, totaling approximately 8,000 sources. To gather metadata for each barrier, the primary task was to identify the headquarters of each news publisher, which we sourced from the infoboxes on Wikipedia. We utilized the Bright Data service [2] to crawl and extract information from Wikipedia infoboxes for over 8,000 news websites. For the economic barrier, we utilized The Legatum Prosperity Index, which evaluates countries based on twelve economic dimensions. Economic disparity was quantified using the Euclidean distance function, and a k-means clustering algorithm was applied to classify countries into 20 economic clusters. The metadata is stored in the barriers database, and the country classifications are available in the GitHub repository [3] (see Table 1 in the appendix).

---

[2] https://brightdata.com/

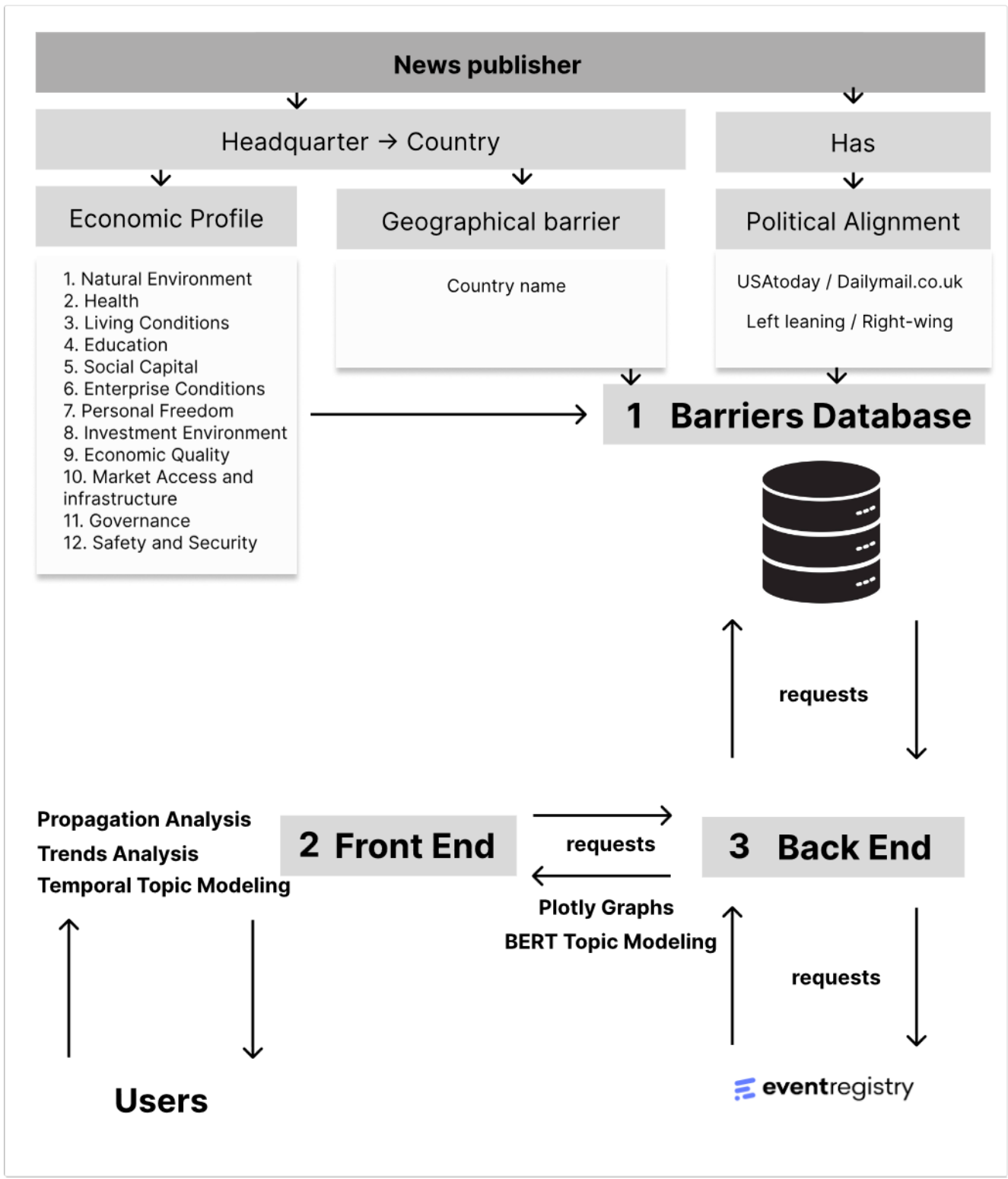

**Figure 3:** The approach for gathering barrier-related metadata and integrating it with other components of the BAR-Analytics platform.

### 4.2. Proposed Analyses for Barrier Differentiation
#### 4.2.1. Propagation Analysis
This analysis examines how news spreads by measuring the similarity between articles using cosine similarity. A dynamic 3D network graph visualizes the propagation, and the Girvan-Newman algorithm is used to detect communities of related articles. Communities with few articles often contain near-identical reports, whereas larger communities indicate evolving narratives (see Section 5.2).

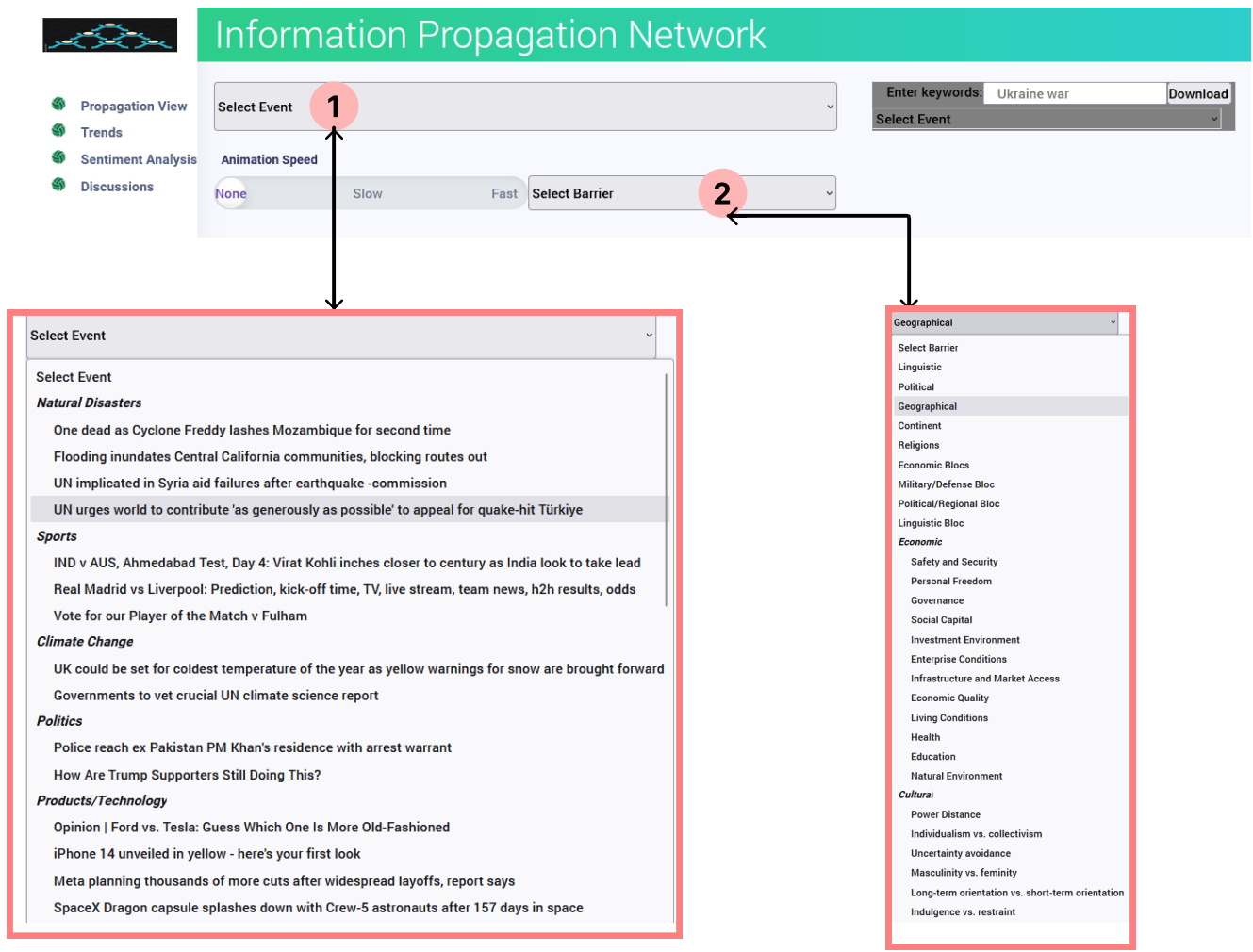

**Figure 4:** Explanation of the BAR-Analytics platform

### 4.2.2. Trends Analysis

Trend analysis focuses on quantifying the frequency of news articles published on selected topics over various time intervals. We represent the findings through line graphs, which help to illustrate the variations in news coverage across different barriers over time (see Section 5.3).

### 4.2.3. Sentiment Analysis

Sentiment analysis evaluates the tone of news articles using VADER, a lexicon-based sentiment analysis tool. Articles are classified as positive, neutral, or negative based on their sentiment scores (see Section 5.4).

### 4.2.4. Topics Analysis

Topic modeling is performed using hierarchical topic modeling through BERTopic to identify key themes within the news corpus. The Cc coherence measure is used to validate topic consistency (see Section 5.5). We represent the frequent topics across barriers in bar charts.

### 4.3. Development of the Web-Based Application

The final aspect of our framework involves the creation of a Flask-based web application, where we designed both the front and back end to connect with the barriers database (see Figure 4). The front end enables users to interact with the BAR-Analytics platform. It is responsible for rendering a variety of data visualizations, including the 3D propagation network, sentiment and trend graphs, as well as bar charts that highlight the topics being discussed in selected news articles (see Figure 5). Interactive graphs are built with web technologies such as HTML, CSS, and JavaScript and Plotly and enable users to explore a comprehensive overview of information related to propagation, sentiment, trends, and topics.

On the backend, the application is powered by Flask, a lightweight and flexible web framework in Python [4]. The backend is tasked with managing requests from the front end, processing the required data, and delivering the necessary information for visualization. It communicates with Event Registry to retrieve pertinent news articles and interacts with the barriers database to extract metadata. This robust backend infrastructure ensures that the data is adequately prepared for presentation, allowing for an efficient and responsive user experience. The source code is available in the GitHub repository [5]. Here is the link to the web application: BAR-Analytics.

---

[4] https://realpython.com/flask-project/

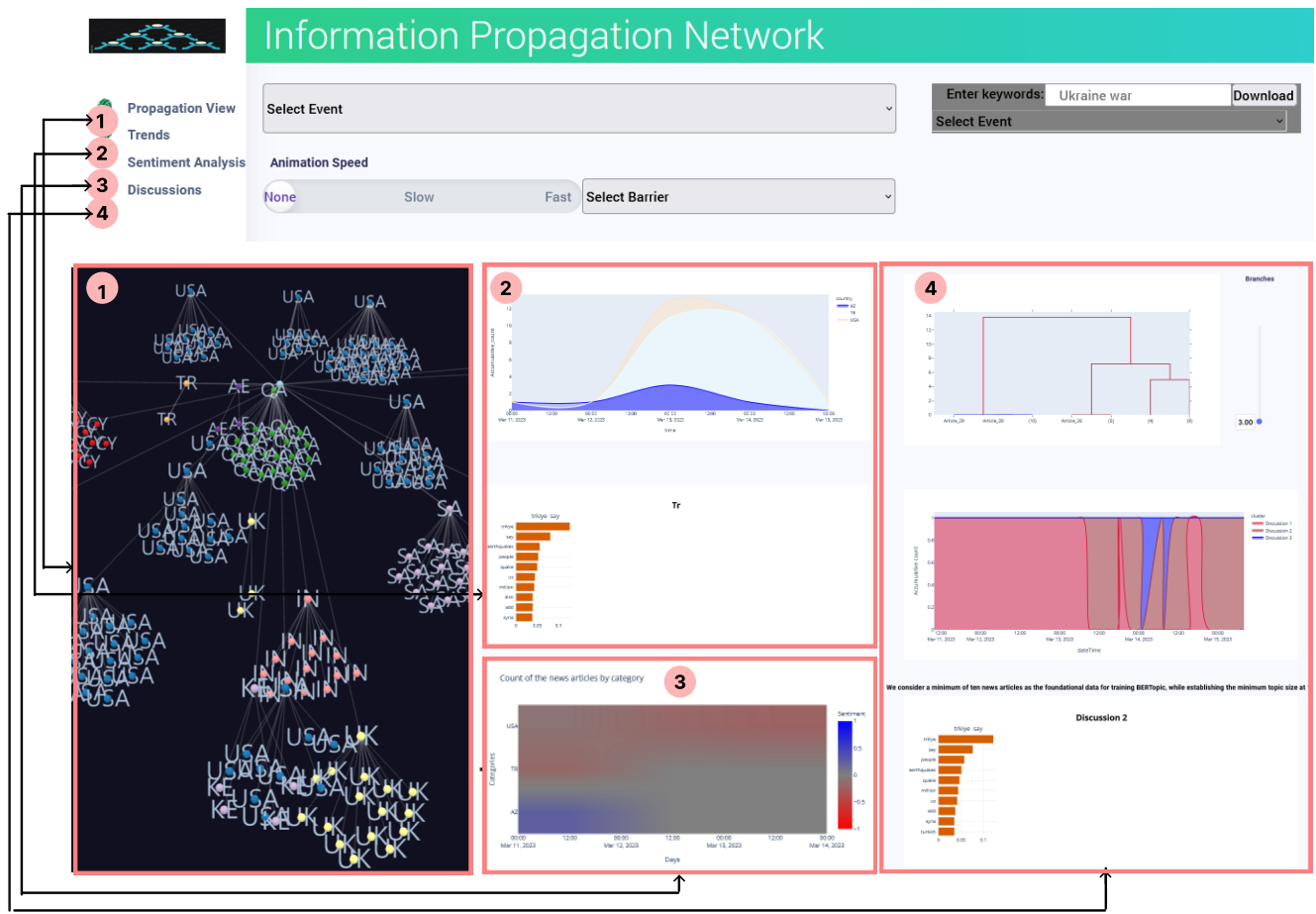

**Figure 5:** Different kind of analysis provided by BAR-Analytics platform

## 5. Use case - Ukraine and Palestine war

To examine the variations in news reporting with the BAR-Analytics platform, we concentrated on two current conflicts: the Russian-Ukrainian war and the Israeli-Palestinian war. We selected these events for two primary reasons. Firstly, both conflicts share similarities, allowing us to explore potential relationships between them through one of our proposed analysis methods. Secondly, these conflicts represent some of the most pressing contemporary issues, providing valuable insights into how news reporting differs across political, economic, and geographical barriers.

---

[5] https://github.com/abdulsittar/BAR-Analytics/

### 5.1. Research Objectives

The primary objective of this study is to validate the developed integrated framework in real-world news reporting setting involving multiple news spreading barriers. It takes the advantage of state-of-the-art tools of natural language processing and web technologies. In this context, we define the following research objectives:

- **RO1. Analyze the differences in propagation, trends, sentiment, and topics across various barriers for both conflicts.**

- **RO2. Determine the similarities and differences among events of the same nature, such as wars.**

### 5.2. Propagation analysis
#### 5.2.1. Israeli-Palestine war

**Economic, political and geographical barriers:** To see the propagation across economic, political, and geographical barriers, we present a network diagram for each barrier. For propagation across the geographical barrier a propagation network of news articles (published during the month of November 2023), structured to highlight their geographical distribution based on the location of the news publishers. It provides an in-depth look at how news content is generated, clustered, and connected globally, allowing us to understand how certain topics circulate across regions. The most striking feature of this network is the way it forms geographical clusters. Articles from the same region or country are grouped tightly together, forming dense clusters. For example, in this visualization, the USA appears to have a vast and densely connected set of news articles, reflecting its high output of media content. These clusters indicate that news publishers in the same region or country tend to cover similar topics, resulting in a significant overlap of content. These "mountain-like" peaks reflect how concentrated the discussion of certain issues becomes within a particular geographical area. This clustering also reveals the dominance of certain countries in global news production. Larger clusters indicate countries with a more substantial presence in the network, meaning they contribute more news articles, which in turn shape the global narrative. The USA's prominence in the visualization shows that it produces a large volume of articles, which form strong internal connections based on similar topics.

Similarly, we present a propagation network of news articles (published during the month of November 2023) across different political alignments (see Figure 13) The dominance of certain clusters, such as those labeled right-wing conservative or center-left, reflects how political orientation influences the reach and impact of news stories. Articles from politically dominant groups tend to have more connections and a wider spread, indicating that these perspectives may be more influential or widely disseminated. At the same time, smaller clusters, such as those labeled independent or pro-Western, suggest that some political perspectives have a more limited reach, with their content remaining confined to smaller, less interconnected networks. This indicates that political minorities or alternative viewpoints may struggle to gain traction in the broader media landscape.

We present an example (see Figure 15) showing the propagation network of news articles across different economic classes (see the Section 4). Each dot on the graph represents a news article, and the lines or edges connecting the dots show how these articles are related based on their content. The network uses the news articles published during the month of November 2023 to reveal patterns in how news stories spread across different economic classes, illustrating the connections between different articles based on common themes or topics. By using this economic classification, the network reveals how different publishers in varying economic environments may approach the same story differently, reflecting either local priorities. The clusters in network indicate groups of articles that are closely related in content or share a common economic context based on the publisher's location. The connections between these clusters, represented by the edges, are formed based on the similarity of the articles, measured through Wikipedia-concept-based similarity.

The structure of the network shows that there are distinct areas where articles are more densely connected. In the upper section of the graph, smaller clusters of nodes suggest localized or specific topics that are only relevant to a certain group of publishers, likely those from a similar economic background. This indicates that some economic issues are more relevant to particular economic regions. Conversely, in the lower part of the graph, larger clusters with more spread-out connections suggest that some news topics are more universally covered or have broader appeal, linking publishers across different economic classes. The clusters and connections between articles reveal the influence of these economic factors on how news spreads, showing patterns of coverage that are shaped by both content similarity and the financial context of the news outlets involved.

### 5.2.2. Russia–Ukraine war

**Economic, political and geographical barriers:** In the Russia-Ukraine war, network diagrams for economic barrier (see figure 16) reveal patterns of news propagation—which regions or economic classes dominate the flow of information and which are more peripheral. A propagation analysis can help determine whether certain economic classes of publishers are driving global conversations or if they are merely reacting to major events reported by larger, wealthier media outlets. The visualization suggests a centralized propagation model, where key news stories spread outward from a few central sources, gradually influencing a broader range of peripheral topics and regions. However, the presence of smaller, independent clusters also implies that localized information flow is occurring independently of the central narrative, driven by regional or economic interests.

The propagation of news articles about the Russia-Ukraine war across various political alignments in April 2023 is presented in Figure 6. Each dot corresponds to a news article, and its position and connections indicate the political orientation of the publishing news outlet, ranging from center-left and center-right to more distinct ideological perspectives such as independent, populist, and conservative. The clusters formed by these articles suggest that news outlets with similar political leanings tend to produce content that shares themes and perspectives, leading to tight groupings of articles based on ideological alignment. For instance, dense clusters of center-left and center-right articles suggest that moderate news outlets on both sides of the spectrum are heavily involved in covering the war. At the same time, the connections between different clusters demonstrate some degree of overlap in the topics covered by ideologically diverse outlets, even though their interpretations of events may differ. Articles from independent outlets form a central hub in the network, indicating that these outlets may offer more neutral coverage, acting as a bridge between polarized perspectives. However, the presence of clusters that lean towards more populist, conservative, or progressive views suggests that there is still a level of polarization in how different political groups frame the war, with some publications potentially focusing on distinct narratives that reflect their ideological leanings. Overall, the network highlights both the interconnectedness of global reporting on the Russia-Ukraine war and the influence of political ideology on how the conflict is portrayed.

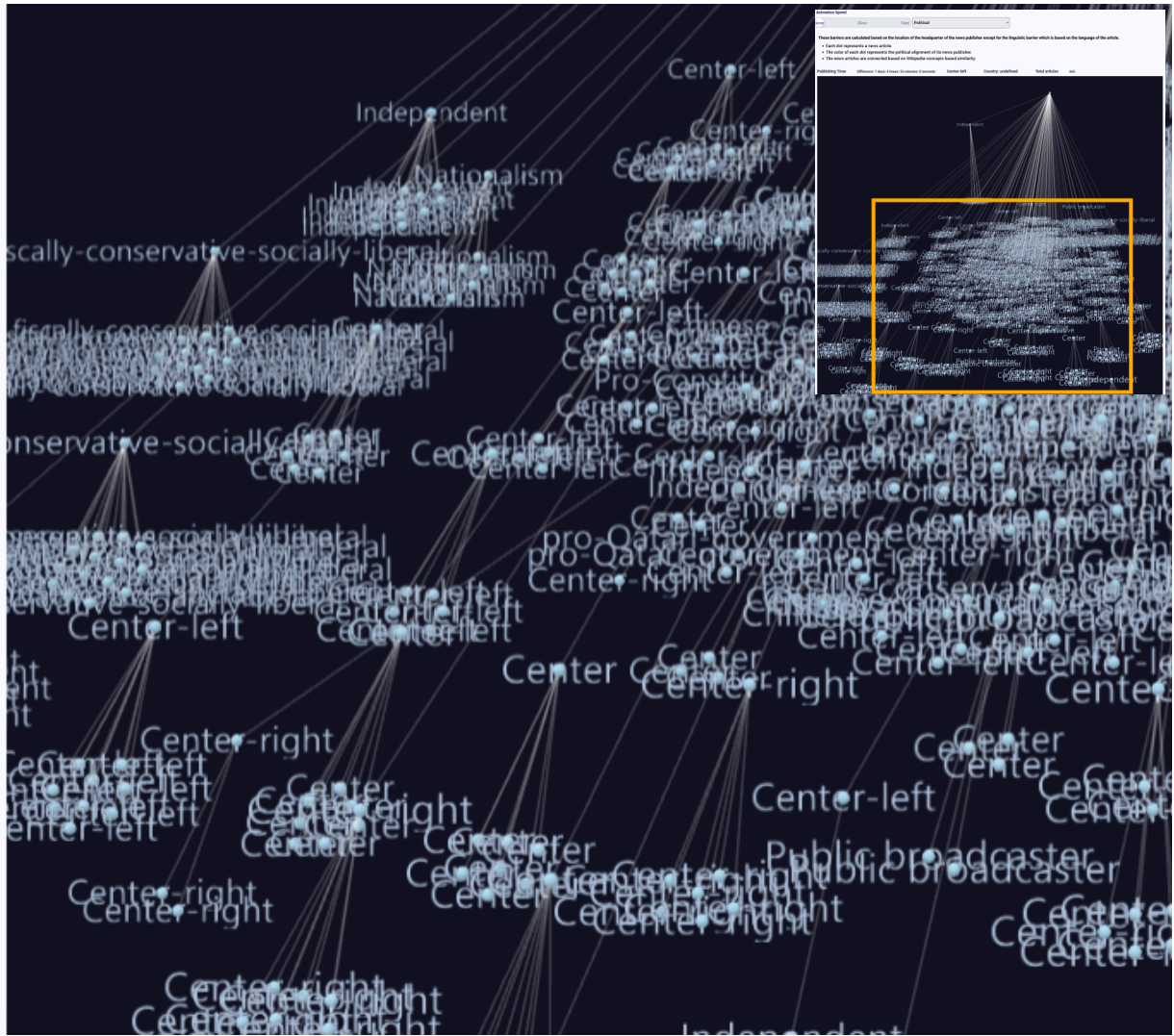

**Figure 6:** The network graph illustrates the propagation network of news articles published in the month of April 2023 related to Russian-Ukraine war. And it represents propagation across political barriers.

In the third visualization (see figure 17), large clusters of articles are visibly linked to prominent regions such as the USA and UK, suggesting that a significant portion of news coverage about the Russia-Ukraine war originates from these locations. The centrality of these regions in the visualization indicates that they play a key role in shaping the global narrative, producing a vast number of articles that are interconnected based on shared topics or content. These clusters are indicative of major media markets in the Western world, where news outlets have extensive resources and influence. The USA, for example, may not only be generating content internally but also influencing the flow of news to other regions, given the connections shown between different clusters. Similarly, the UK's central role likely reflects its position as a major hub for English-language journalism. The lines connecting the clusters represent the pathways through which news propagates across geographical boundaries. For instance, articles produced in the USA are connected to articles in the UK and other parts of the world, showing how news travels and is referenced or shared internationally. These connections suggest that news articles are not confined to their region of origin but rather flow across borders, influencing the reporting in other countries.

### 5.3. Trends analysis
#### 5.3.1. Israeli-Palestine war

**Political, economic, and geographical barriers** Understanding how the news propagate, we were next interested their occurrence over the barriers (see Figure 7). For the political barrier, news articles published in November 2023 were categorized by their political alignments (Figure 7b). The peaks in the graph represent spikes in political news coverage. These spikes likely correspond to significant political events or controversies, which prompted increased media activity. For example, around November 7, there is a noticeable peak in the graph, followed by another around November 20, and yet another towards the end of the month, around November 27. These peaks are mostly dominated by center-left media, as shown by the large light blue areas, indicating that much of the political reporting during these periods came from outlets with a center-left orientation. Other political alignments, such as center-right liberal and conservative, also contributed to the coverage, but their impact appears smaller in comparison. The graph effectively illustrates the distribution of political discourse across different ideological lines. By comparing the areas of color, it becomes clear which political voices were most prominent at different times. The dominance of center-left media coverage suggests that this perspective had a strong influence on the political conversation during November 2023. However, the presence of other alignments, such as conservative and independent outlets, also suggests a diversity of viewpoints contributing to the broader political discourse, thus leading to a comprehensive view of how political news was reported across various ideological spectrums throughout November 2023.

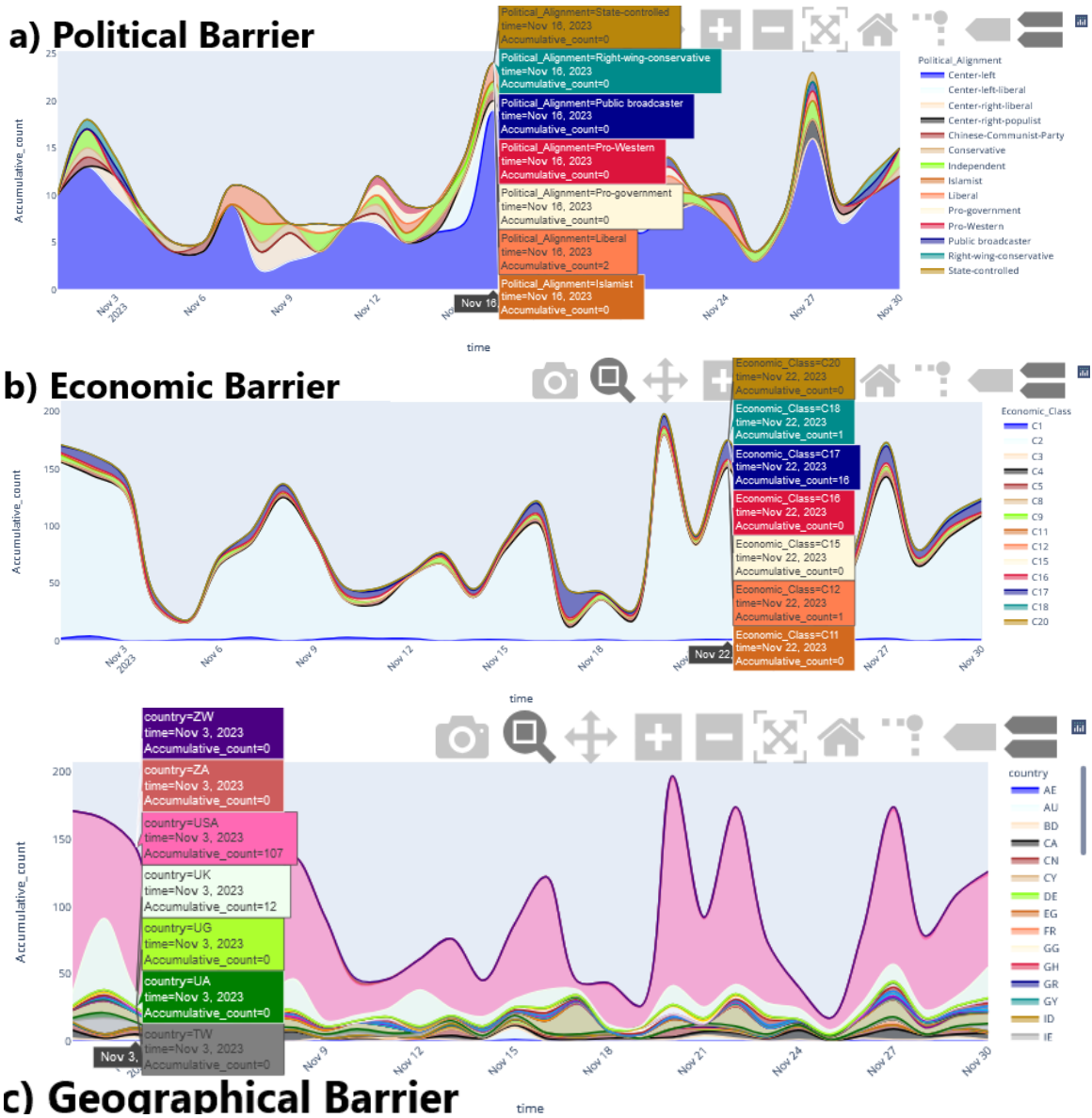

**Figure 7:** The line graphs illustrate the trends of news articles published in the month of November 2023 related to Israeli-Palestine war. And each graph (from top to bottom) represents trends across political, economic, and geographical barriers respectively.

The trends of news articles related to the Israel-Palestine war, segmented by different economic classes (C1, C2, etc.) for November 2023 are shown in Figure 7a. Each line corresponds to a particular economic class, indicating the level of coverage or attention given to the conflict in that class, shown cumulatively. Peaks in the graph suggest heightened media attention during certain periods, potentially reflecting major developments in the war. The pattern shows that most economic classes experience spikes in news coverage around the same key dates, such as November 6, 10, and 27, suggesting synchronized reactions to significant events. The variation in cumulative counts across classes highlights differences in the volume of coverage or the importance assigned to the conflict, with some classes (e.g., C17, C18) showing higher counts, indicating a broader focus in those

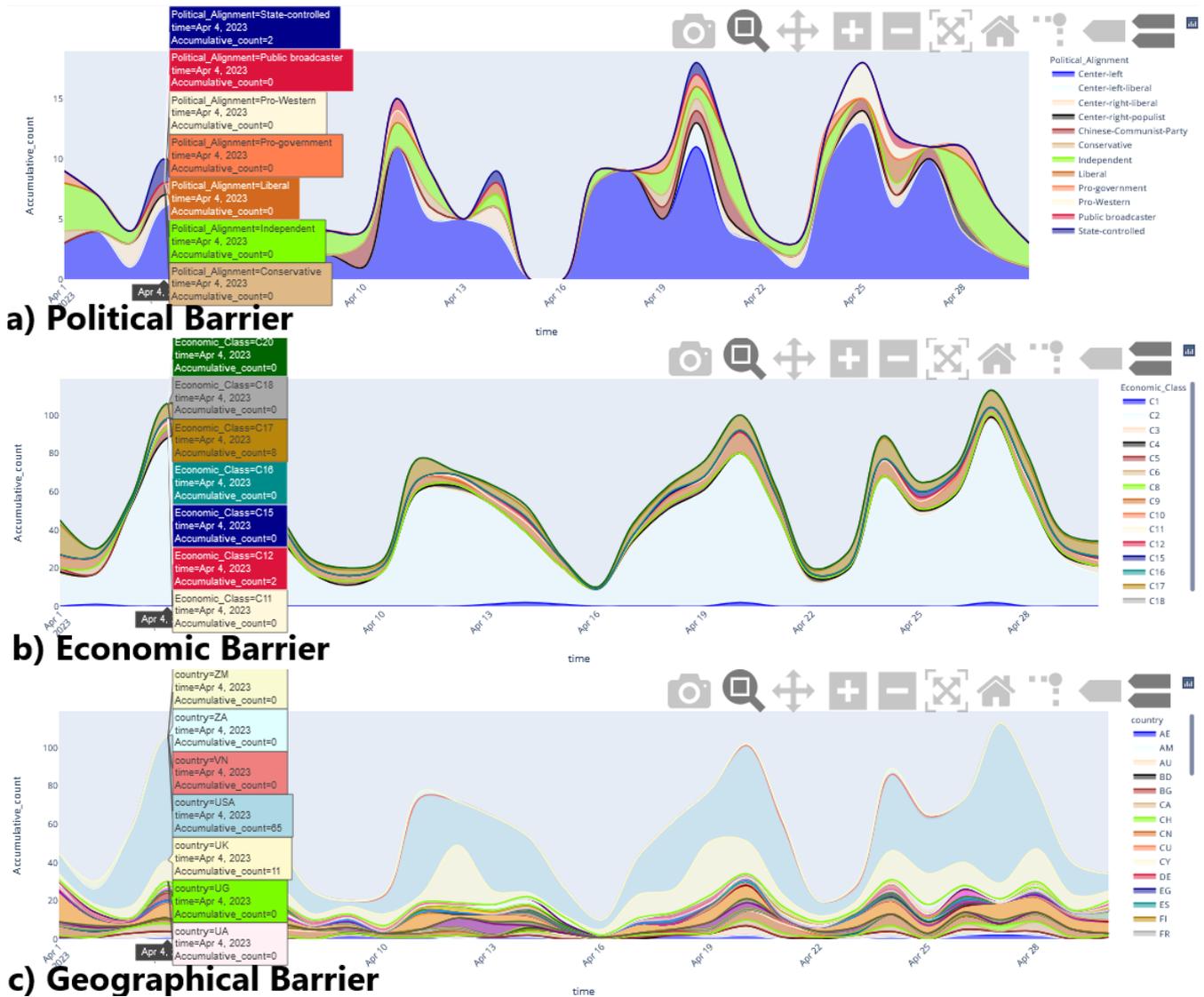

**a) Political Barrier**

**b) Economic Barrier**

**c) Geographical Barrier**

**Figure 8:** The line graphs illustrate the trends of news articles published in the month of April 2023 related to Russia–Ukraine war. Each graph represents trends across political, economic, and geographical barriers respectively.

economic groups (e.g countries such as Botswana, El Salvador, India, Indonesia, Mexico, see Table 1 for more). This likely reflects varying levels of interest or impact across different economic groups, possibly due to the geopolitical or financial relevance of the war to each class.

To study how geographical barrier affects trends of news, news' publisher headquarters countries were extracted (Figure 7c). The line graph shows the cumulative number of news articles published over time, categorized by country for November 2023. Different colors represent various countries, with larger areas indicating a higher volume of news articles. For example, the pink area corresponding to the USA exhibits significant peaks around mid and late November, indicating a surge in article counts during these periods. Other countries like South Africa (ZA) and Zimbabwe (ZW) contribute smaller but noticeable amounts to the total.

### 5.3.2. Russia–Ukraine war

**Political, economic and geographical barriers** For the political barrier, news were segmented by the political alignment of different news publishers for April 2023 (Figure 8b). The various colored layers correspond to the political alignment categories, such as "State-controlled," "Pro-government," "Independent," "Liberal," and others. A noticeable pattern is the periodic surge in news publications, particularly around April 11, April 20, and April 25, where the activity spikes. "State-controlled" media, represented by the thick blue section, consistently dominates the landscape, contributing the most articles, followed by other categories like "Public broadcaster" and "Pro-government" (seen in lighter colors). This trend indicates that state-influenced or controlled media had a larger presence in reporting on the Russia-Ukraine war, while other categories, such as "Conservative" or "Liberal," contributed less frequently but were still present throughout the month. The periodic peaks could correspond to significant events or escalations in the conflict, which drove increased media coverage across all political alignments.

Economic barrier related to the Ukraine-Russia war is presented in Figure 8a,, categorized by different economic classes (C1, C2, etc.), over the month of April 2023. The multiple economic classes represented by colored lines show fluctuations in media coverage, with several peaks occurring on specific days, indicating key events in the conflict that drew significant attention across all economic classes. The peaks around April 4, April 10, and later dates suggest major developments in the Ukraine-Russia war that were widely covered. The variation in the magnitude of these peaks between classes (e.g., C18, C20, and others) indicates that certain economic groups (see Table 1) were more actively discussing or impacted by the war. This could reflect varying levels of geopolitical, financial, or social interest in the conflict depending on the economic standing of these classes, with some showing higher concern or engagement compared to others.

For the geographical barrier, trend of news articles related to the Ukraine-Russia war across different countries throughout April 2023 are shown in Figure 8c. Different countries, represented by colored areas, demonstrate varying levels of media coverage on the topic. The larger areas like those associated with the USA, UK, and Ukraine (UA) indicate that these countries had significantly higher media attention focused on the conflict during specific time periods, such as around April 4, April 10, and April 20. This suggests key events or developments in the war during these peaks that garnered widespread international media attention. The smaller areas, representing countries such as Zambia (ZM) and Vietnam (VN), show limited coverage or interest, which could be due to lower geopolitical involvement or national media priorities. The recurring peaks across almost all countries suggest that certain moments in the Ukraine-Russia war prompted synchronized media focus worldwide, with noticeable fluctuations between countries, indicating varying degrees of engagement with the conflict based on regional or national interests. Overall, the graph reveals global media trends and highlights the countries most actively reporting on the war.

## 5.4. Sentiment analysis

Analyzing the sentiments in news articles can help to see the changes in sentiments over time and to make comparison of sentiments across different countries.

### 5.4.1. Israeli-Palestine war

**Political, economic, and geographical barriers** We first analyzed the sentiment trends related to news articles on the Israel-Palestine war ( see Figure 9a). For the political barrier, news articles published throughout November 2023 were segmented by political alignments. The analysis is depicted as a heatmap, with a color scale indicating the sentiment of articles published by various political categories over time. Each day represents the average sentiment of news articles published by different political categories. The heatmap uses a gradient of colors to represent sentiment: blue color indicates positive sentiment, red color indicates negative sentiment, and gray color represents neutral sentiment or a lack of significant sentiment. The color scale on the right quantifies this: darker shades of blue suggest stronger positivity, while darker shades of red indicate deeper negativity. Neutral sentiment, shown in gray, represents moments where the news coverage did not lean heavily towards either a positive or negative tone.

Taking a closer look at the heatmap, patterns begin to emerge. For instance, there are pockets of negative sentiment (red) around specific days, such as November 12 and November 24, where certain political alignments like "Right-wing-conservative" and "Pro-government" display stronger negative tones. This suggests that news related to particular political developments during these periods was framed unfavorably by these outlets. Conversely, positive sentiment (blue) is visible in other areas, such as the "Center-left" around November 9 and November 27, indicating more favorable or optimistic coverage from those political perspectives on those days. Similarly, positive sentiment is almost visible for Pro-government and Chinese Communist Party news outlets. The fact that both positive and negative sentiments are visible throughout the heatmap suggests a polarized media landscape, with different ideological groups interpreting and presenting political events in contrasting ways, which may influence public perception based on the media they consume. Overall, the heatmap provides a detailed view of how political events were framed by different media outlets, revealing how different perspectives can lead to starkly different representations of the same events.

To understand the economical barriers, news were categorized by economic classes (C1, C2, etc.) and analyzed for November 2023 (see Figure 9b). Throughout the period, we observe alternating streaks of positive and negative sentiment for various economic classes. For example, in the first week of November (around Nov 5), negative sentiment (red) is prominent in classes like C2, C3, and C5, while positive sentiment (blue) appears more in C1 and C9. As November progresses, there is a visible shift with increased neutral (gray) and positive coverage around November 15, followed by more intensified negative sentiment after November 19. This suggests that different economic classes may perceive or report on the Israel-Palestine war differently, possibly reflecting their specific economic interests or geopolitical positions.

To understand how geographical barriers portray Israel-Palestine war, news were categorized based on the countries (see Figure 9c) during November 2023. The colors represent sentiment polarity: red for negative sentiment, blue for positive sentiment, and gray for neutral or less intense sentiment. Throughout the month, different countries exhibit varying sentiment patterns. For instance, countries like Israel (IL), USA, and some Middle Eastern nations (e.g., UAE, SA for Saudi Arabia) display more frequent negative sentiment (red) across multiple dates, particularly after November 12. In contrast, countries like New Zealand (NZ), Pakistan (PK), and Switzerland (CH) have occasional pockets of positive sentiment (blue), although the general tone across most countries remains relatively neutral (gray). This suggests that geographical location plays a role in shaping the sentiment of news coverage, likely influenced by political affiliations, regional interests, or the direct impact of the conflict on each country.

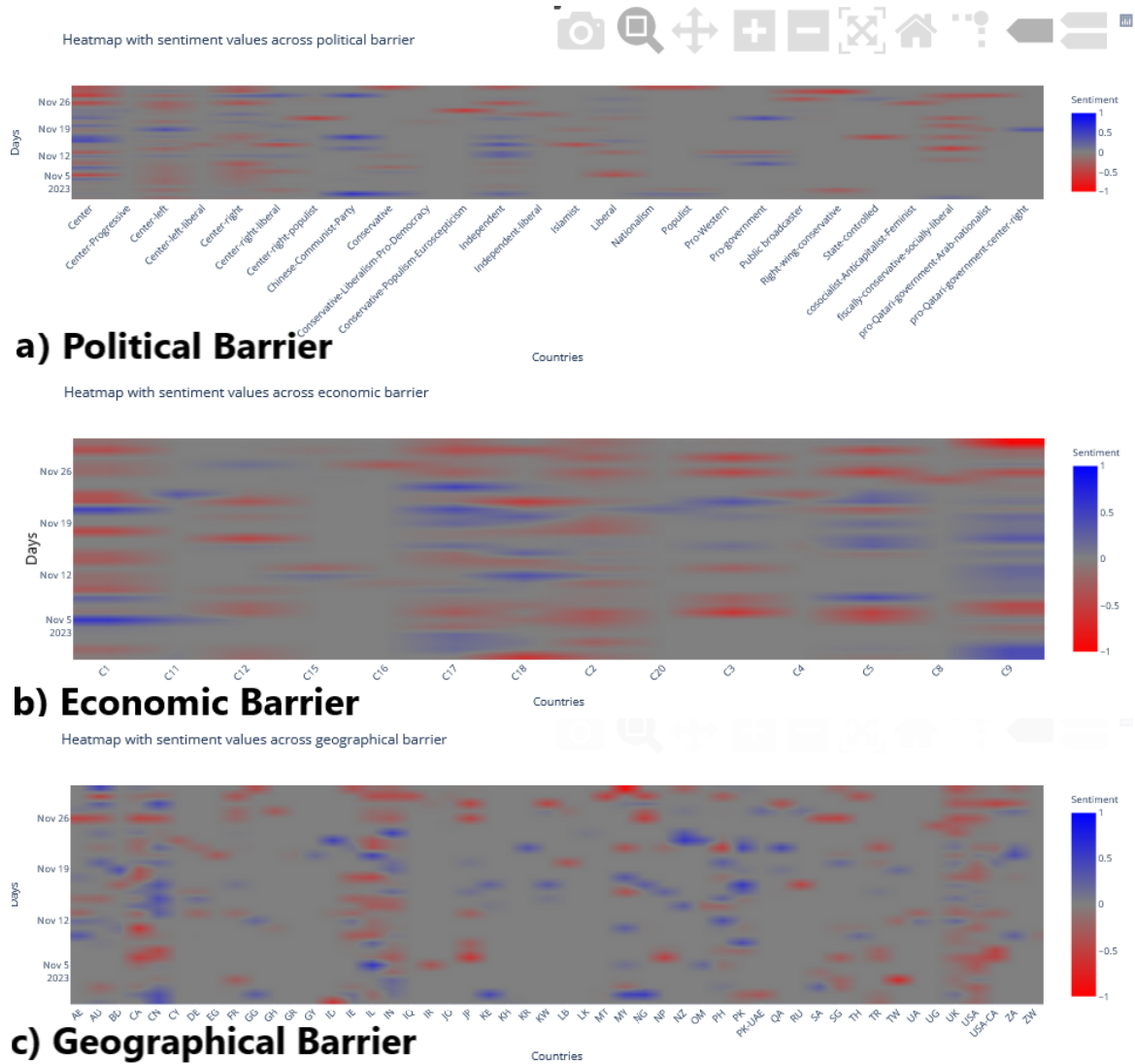

**Figure 9:** The heat charts illustrate the sentiments of news articles published in the month of November 2023 related to Israeli-Palestine war. Each graph represents sentiments across political (a), economic (b), and geographical (c) barriers respectively.

### 5.4.2. Russia–Ukraine war

**Political, economic, and geographical barriers** To analyze the sentiment of news articles related to the Russia-Ukraine war published throughout April 2023, we categorized the political alignment of the news publishers (see Figure 10a). The vertical axis represents the days of April, while the horizontal axis lists various political affiliations or ideological stances of the news sources, such as "Center-left liberal," "Conservative," "Pro-Western," "Public broadcaster," and more nuanced categories like "Pro-Establishment," "Nationalist," and "State-controlled." The color scale reflects sentiment: blue indicates positive sentiment, red shows negative sentiment, and grey represents neutral or absent sentiment.

The figure shows significant variation in sentiment across political alignments and over time. For instance, left-leaning or liberal sources seem to express a mixture of positive and neutral sentiments (more blue and grey areas) regarding the conflict, whereas more conservative or pro-government publications exhibit periods of negative sentiment (red areas), particularly mid-month. Certain categories like "State-controlled" and "Pro-Russian" tend to show more negative sentiment or neutrality, whereas "Pro-Western" or "Public broadcaster" sources lean toward a more neutral or slightly positive stance. The color distribution highlights how different political ideologies influence media coverage and sentiment on the Russia-Ukraine war, with some sources aligning more favorably or critically depending on their

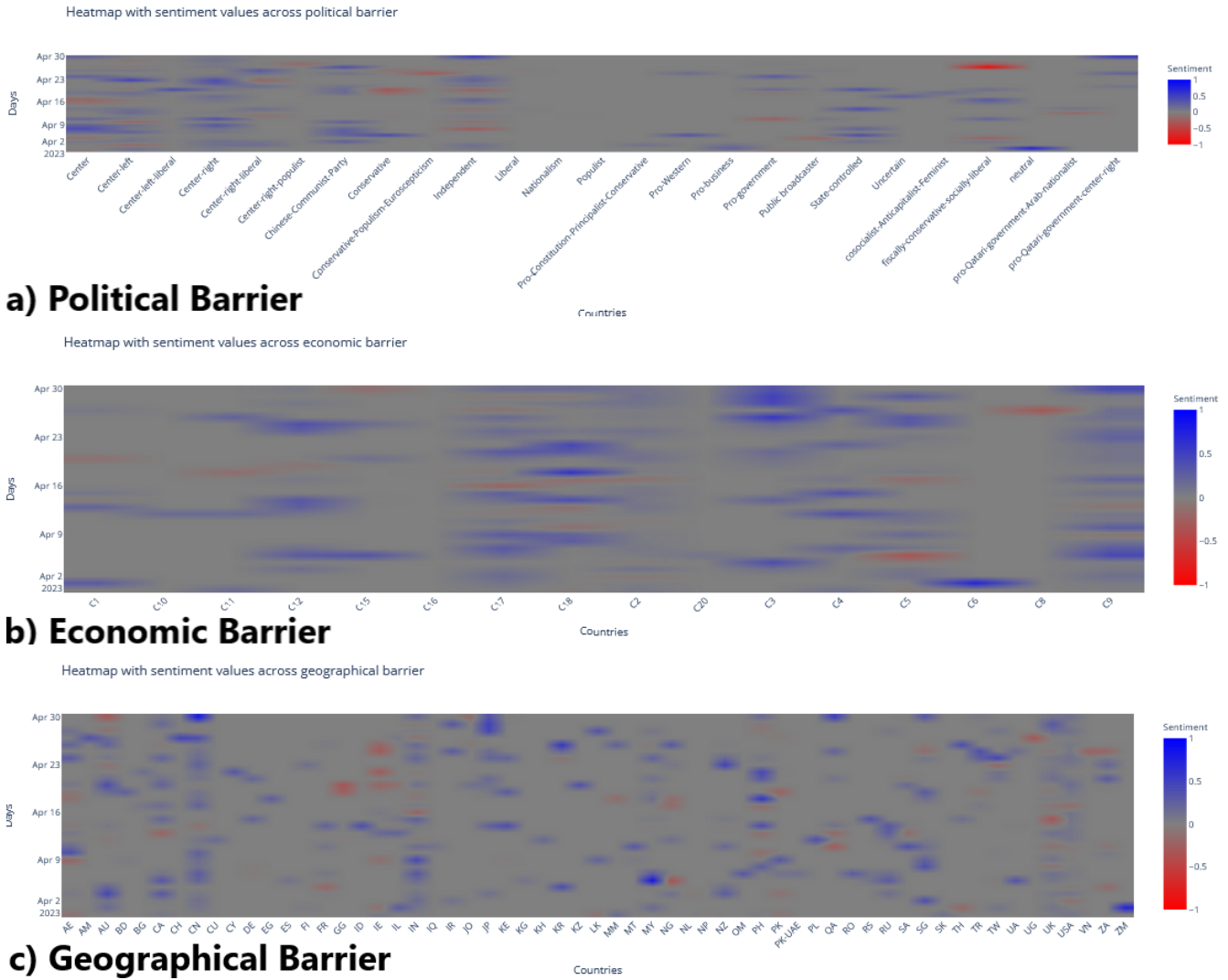

Heatmap with sentiment values across political barrier

## a) Political Barrier

Heatmap with sentiment values across economic barrier

## b) Economic Barrier

Heatmap with sentiment values across geographical barrier

## c) Geographical Barrier

**Figure 10:** The heat charts illustrate the sentiments of news articles published in the month of April 2023 related to Ukraine-Russian war. And each graph (from top to bottom) represents sentiments across political, economic, and geographical barriers respectively.

political orientation.

For the economical barriers, heatmap (see Figure 10b) visually represents the sentiment of news articles about the Russia-Ukraine war across various economic classes (denoted as C1, C2, C3, etc.) for the month of April 2023. The vertical axis represents time, broken down by days (from April 2 to April 30), while the horizontal axis lists the different economic classifications . Each class (C1, C2, etc.) corresponds to a type of economy or country grouping, though the specific definitions of these classes are not visible here. The color gradient represents the sentiment, with blue indicating more positive sentiment, red showing more negative sentiment, and grey areas likely representing neutral or insignificant sentiment. From the figure, it can be observed that sentiment varied throughout the month and across different economic classifications. For instance, certain economic groups (like C2, C6, and C10) show more concentrated positive sentiment (blue), while others display brief episodes of negative sentiment (red). The spread of colors suggests that the reactions to the war in the media were not uniform, possibly influenced by each economy's direct or indirect involvement in or proximity to the conflict. Additionally, the presence of neutral or low-intensity sentiment (grey) implies either less reporting or a lack of strong emotional tone in those regions or during those periods. For the geographical barrier, heatmap illustrates the sentiment of news articles published in April 2023 concerning the Russia-Ukraine war across different countries (see Figure 10c). The vertical axis represents the days of the month, ranging from April 2 to April 30, while the horizontal axis lists the countries (abbreviated by their ISO country codes) such as AU (Australia), BR (Brazil), CN (China), and RU (Russia). The color scale indicates the sentiment intensity: blue represents positive sentiment, red signifies negative sentiment, and grey suggests neutral or negligible sentiment.

The distribution of blue and red regions across countries shows variations in how news articles portrayed the Russia-Ukraine war globally. For example, countries like AU, BR, and US display more prominent positive sentiment (blue patches), while others, such as RU and some European countries, show sporadic negative sentiment (red patches), particularly around mid-April. Many countries show grey, indicating a neutral stance or lack of strong sentiment on certain days. This heatmap reflects the diverse geopolitical reactions and media coverage patterns, with certain countries experiencing more pronounced fluctuations in sentiment over time, possibly influenced by their geopolitical positions or interests in the conflict.

## 5.5. Temporal Topic Modeling

To filter the dataset, topics were manually selected based on hierarchical clustering. For the Israeli-Palestine war, topics were aggregated across all economic classes, political orientations, and countries due to data sparsity, which limited direct comparisons between intermediate classes. To ensure the quality and relevance of the topics, we validated our results using various metrics, including topic coherence and topic diversity, which help assess the internal consistency and coverage of the identified topics, ensuring their robustness and relevance.

### 5.5.1. Israeli-Palestine war

**Economic Barrier:** Graph (a) in Figure 11 illustrates the frequency of economic topics over time as they propagate across the economic barrier. This area chart represents various economic factors such as currency stability, trade dynamics, inflation, market conditions, unemployment rates, and budgetary concerns, indicating how economic discussion evolve in response to cross-border trade tensions, and resource accessibility. Peaks in the graph suggest periods in reaction to policy changes, sanctions, or shifts in the international economic landscape that directly impact the economic connectivity across the economic barrier. Cumulatively, agreements and trade were the most predominant topics. General topics such as market, inflation, interest and stocks were stable throughout the whole period, others only for a shorter period at the beginning of the war (e.g., unemployment, sanctions, budgets).

**Political Barrier:** Graph (b) in Figure 11 focuses on the propagation of political topics across the political barrier. This chart captures discussions around political dimensions, such as border disputes, judicial issues, military strategies, peacekeeping efforts, diplomatic engagements, sanctions, and human rights concerns. Fluctuations in the frequency of these topics likely correspond to critical political events, agreements, or escalations in political tension between the involved parties. Peaks in this graph may represent intensified diplomatic actions, political crises, or significant announcements regarding changes that affect both sides of the barrier. As the current Israeli-Palestine war started in October 2023, topics such as diplomacy and sanctions had the peak values between Oct-Dec 2023. Human rights violations topic increased after Nov 2023 and had a stable consistency throughout the whole recorded period, whereas military strategy dropped after initial peak but climbed up again towards the second half of the first year.

**Geographical Barrier:** Graph (c) in Figure 11 represents the frequency of geographical topics as they propagate across the geographical barrier. Here, the topics include infrastructure damage, refugee movements, supply chain disruptions, resource access, and evacuation activities. These elements reflect physical and logistical concerns directly impacting individuals and communities across the region. Peaks in the geographical topic frequency could correspond to periods of intensified conflict, infrastructure damage, or humanitarian crises that drive discussions on logistical barriers and accessibility across geographical boundaries. Topic of Peace-making missions peaked in Nov 2023 and declined towards the second half of the recorded period. Border disputes, Military strategy, Refuge movements, and Supply chain disruptions had several fluctuations through the year, whereas Information war, Infrastructure damage and Sanctioned regions were stable throughout the whole period.

### 5.5.2. Russia–Ukraine war

Since the dataset for the Russia-Ukraine war was larger compared to Israeli-Palestine war, there was enough data to quantify the topics by economic class, political orientation of the news outlet or the country, where the headquarters of the news outlet are located.

**Economic Barrier:** Graph (a) in Figure 12 tracks topics over time as they spread across an economic barrier, reflecting interactions with distinct country groups labeled as C2, C5, C9, and C17. These groups, listed in the table 1, range from Western countries (e.g., the U.S., U.K., and European Union nations in C2) to Middle Eastern and Asian countries (C9), and include countries from South Asia and Africa (C5 and C17). Key topics within this chart, such as sanctions, trade, and technology, rise and fall in frequency, likely in response to sanctions, currency fluctuations, and market disruptions arising from the Russia-Ukraine war. Each country group's response varies; for instance, C2 countries frequently impose sanctions, impacting global markets, while C9 countries may engage more in trade-related dialogues with Russia. This diversity in reactions highlights the complex global economic landscape shaped by varying regional priorities and alliances within the conflict. Topics Agreements and Trade were uniformly represented in all classes. Sanctions, market and Exports were more common in C9 and C17. Technology was almost uniquely represented in C9.

**Political Barrier:** Graph (b) in Figure 12 presents political topics and their propagation across a political barrier, where discussions span issues such as diplomacy, sanctions, election interference, and military alliances. These topics likely fluctuate in response to key events in the Russia-Ukraine war, such as shifts in military alliances, peace negotiations, or election-related disinformation campaigns. As the graph indicates, political topics often surge in correlation with specific geopolitical actions or policy announcements, reflecting how political barriers contribute to polarized discourse between different countries and alliances. The persistence of these topics across time suggests that political concerns, including interference and alliance formation, are central to understanding how countries perceive and respond to the conflict's implications on governance and sovereignty. To analyze the political barriers, news outlets were clustered based on the self-declared political orientation. Topic of Diplomacy was universally represented along all political orientations (left, centre, right and independent), followed by Sanctions and Military alliance in the first three. Interestingly, Election interference was a prominent topic for left and centre-oriented news outlets in the period from April 2023 on, which corresponds to the period of preparation for presidential elections in Ukraine that in the end did not take place in spring of 2024.

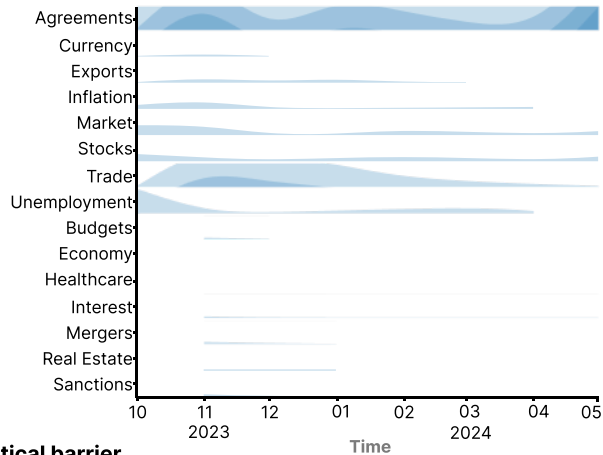

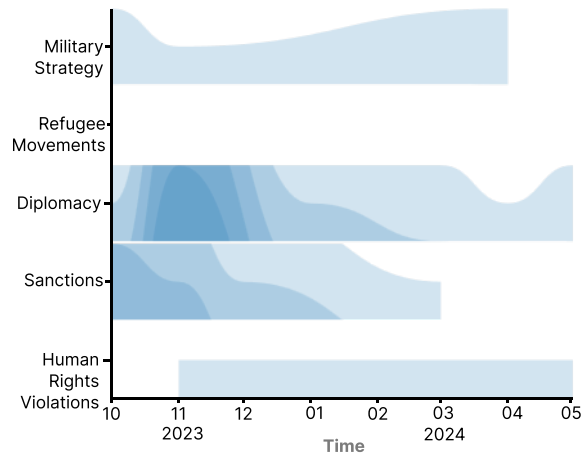

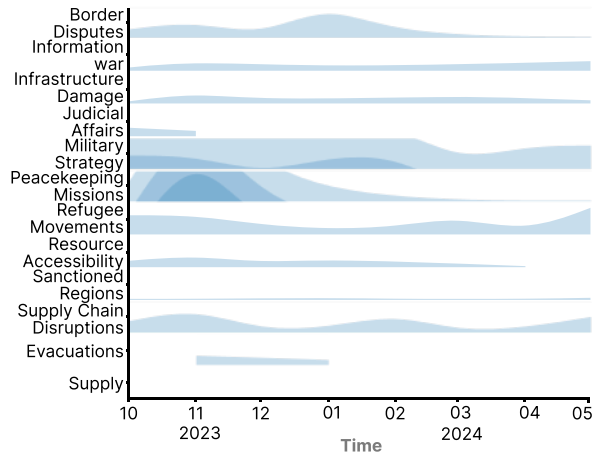

**Figure 11:** The area charts show the frequency of topics over time related to Israel-Palestine war. And each graph represents propagation of barrier-related topics across a barrier. Figure a, b, and c (economical barrier, political barrier, and geographical barrier) presents the propagation of economic topics over economic barrier, political topics over political barrier, and geographical topics across geographical barrier respectively.

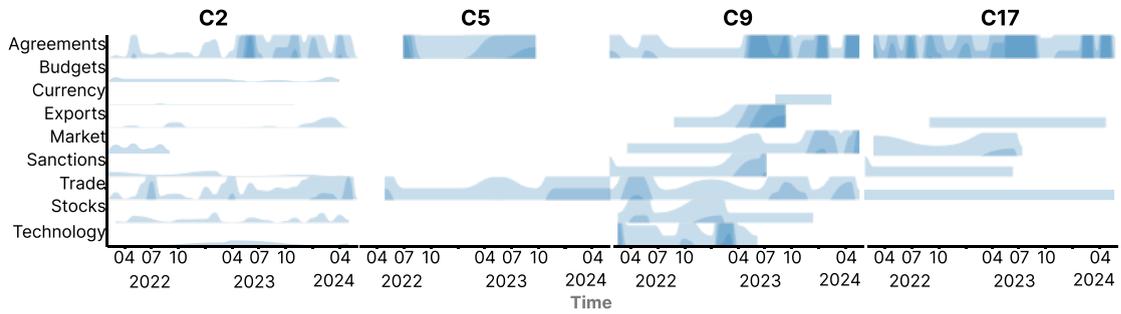

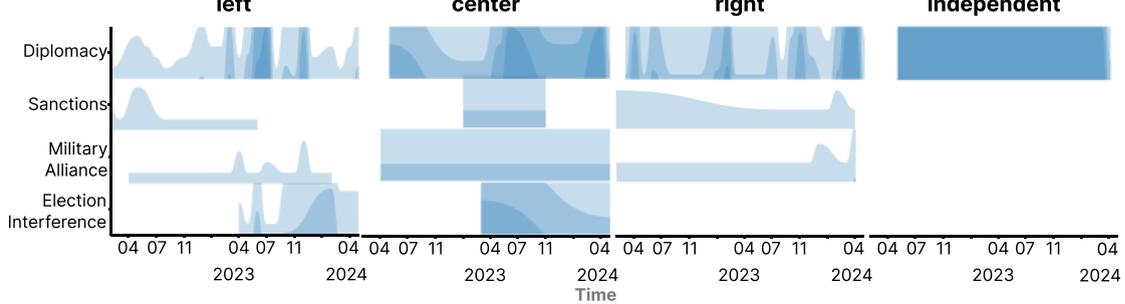

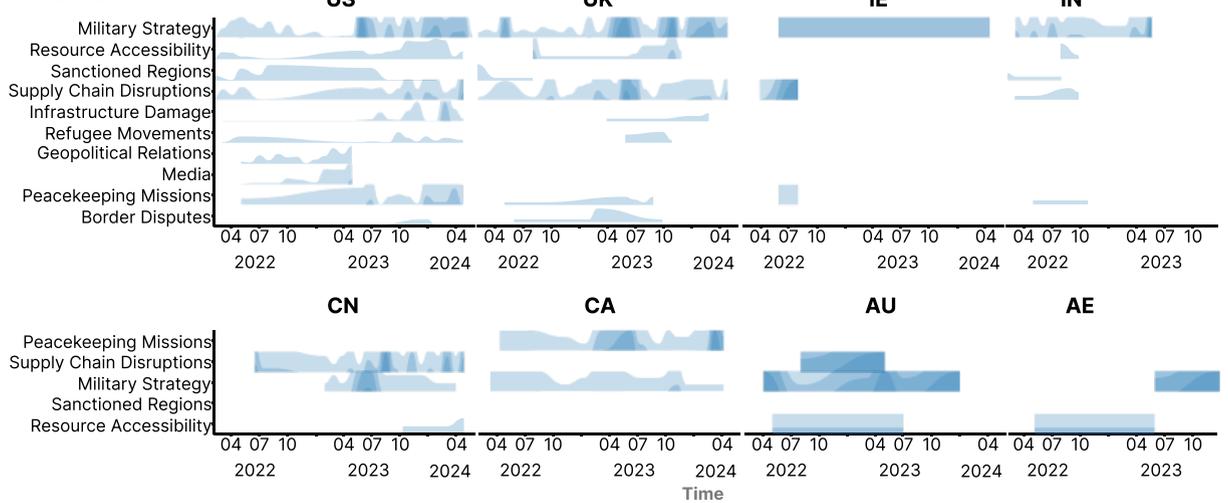

**Figure 12:** The area charts show the frequency of topics over time related to Russia-Ukraine war. And each graph represents propagation of barrier-related topics across a barrier. Figure a, b, and c (economical barrier, political barrier, and geographical barrier) presents the propagation of economic topics over economic barrier, political topics over political barrier, and geographical topics across geographical barrier respectively.

**Geographical Barrier:** Graph (c) in Figure 12 focuses on geographical topics like military strategy, refugee movements, border disputes, and supply chain disruptions, reflecting how physical aspects of the conflict shape and restrict cross-border discussions and humanitarian responses. Peaks in this graph are likely aligned with intense military confrontations, infrastructure damage, and waves of refugee displacement, which heavily influence both local and international discourse on logistics and peacekeeping. Countries close to or directly affected by the geographical aspects of the Russia-Ukraine war, such as those in Eastern

Europe and bordering regions, are particularly impacted by these issues. This graph illustrates the tangible humanitarian and logistical effects of the conflict, showing how geographical barriers restrict resource accessibility and intensify discussions around international peacekeeping and crisis management. In news on Russia-Ukraine war, only 8 exhibited enough data to be visualized: United states, United Kingdom, Ireland, India, China, Canada, Australia, United Arab Emirates. Topic of Military strategy was omnipresent. Interestingly, Peace-keeping missions were present only in the Western countries (US, UK, IE, IN, CA), but not others. US had the most topics, but that is also due to the fact the pool of news was larger.

## 6. Results

In this section, we present an analysis of news propagation, trends, sentiment, and temporal topic modeling for both the Israeli-Palestinian and Russia-Ukraine wars. We explore how economic, political, and geographical factors shape news coverage and how these factors evolve over time, providing a comprehensive overview of the media dynamics in both conflicts.

News propagation during both the Israeli-Palestinian and Russia-Ukraine wars is heavily influenced by economic, political, and geographical factors. In both cases, the USA plays a central role as a hub for global media dissemination. The Israeli-Palestinian conflict sees right-wing and center-left media forming dominant clusters, with wealthier media outlets driving broader coverage. Similarly, the Russia-Ukraine war sees the USA and UK as primary sources of global narratives, with state-controlled media in Russia and other countries forming more isolated clusters. In both conflicts, political alignment and economic power dictate how news spreads, with dominant economic regions shaping international media flows and alternative viewpoints facing greater challenges to gain widespread attention.

Trends in news coverage of both wars show how significant events trigger peaks in reporting, with political and economic factors driving the intensity and timing of coverage. In the Israeli-Palestinian conflict, synchronized coverage peaks correspond to key moments, particularly around early November. Economically, countries like India and Mexico exhibit higher coverage, indicating geopolitical interests. The Russia-Ukraine war also sees fluctuations in media attention, especially around major military and political developments, with countries more economically tied to Russia providing more consistent coverage. The political orientation of media outlets further influences coverage, with center-left outlets playing a dominant role in both conflicts, shaping the broader ideological discourse.

Sentiment analysis reveals how media sentiment is shaped by political and economic contexts in both wars. The Israeli-Palestinian conflict shows a clear divide, with right-wing outlets framing events negatively and center-left outlets presenting a more positive tone, particularly around key moments in November. Similarly, the Russia-Ukraine war exhibits polarized sentiment, with left-leaning and liberal media outlets showing positive or neutral tones, while conservative outlets reflect negative sentiment, especially in Russia. Economic factors also influence sentiment, with wealthier nations generally displaying more positive perspectives, while smaller economic classes lean towards neutral or negative tones. Geographical differences further highlight how sentiment varies, with countries more directly involved in the conflicts exhibiting stronger emotional biases.

Temporal topic modeling reveals how the discussion around economic, political, and geographical topics evolves over time in both conflicts. In the Israeli-Palestinian conflict, early discussions focus on economic concerns such as sanctions and trade, but later shift towards humanitarian issues, diplomacy, and military strategies, with peaks during diplomatic efforts in November 2023. Similarly, the Russia-Ukraine war sees economic topics like trade and sanctions consistently discussed, with political topics such as diplomacy and election interference becoming more prominent as the war progresses. Geographically, the focus on military strategy remains constant in both wars, while topics like peacekeeping and refugee movements fluctuate, reflecting the shifting priorities and international responses to each conflict.

## 7. Analysis and Discussion

This case study, using the BAR-analytical platform as a primary methodology, provides a nuanced view of topic propagation across economic, political, and geographical barriers within the Israel-Palestine and Russia-Ukraine wars, emphasizing the distinctive impact each barrier imposes. The BAR-analytical platform allows for the systematic tracking of topic frequency, revealing patterns that align with real-world events and policy shifts, such as the economic disruptions from sanctions (Graph (a) Figure 11), political escalations impacting diplomatic efforts (Graph (b) Figure 11), and humanitarian concerns associated with refugee movements and infrastructure damage (Graph (c) Figure 11). These insights reflect the complex interplay among economic, political, and geographical dimensions of the conflict, underscoring the need for holistic approaches to address such multi-dimensional issues (Jones, 2016), (Khalidi, 2020). Similar to other platforms like GDELT, which analyzes global news for event data (Rosmer, 2013), (Leetaru and Schrodt, 2013), and Media Cloud, which maps topic trajectories in news coverage (Roberts, 2016), the BAR-analytical platform demonstrates the potential of systematic media analysis in identifying cross-border narrative shifts and barriers. This capability, as shown in the BAR platform's analysis, reveals how each dimension reinforces the others, suggesting that conflict resolution approaches should consider economic, political, and geographical factors together rather than in isolation.

By systematically examining fluctuations in topic frequencies related to Russia-Ukraine, the BAR platform reveals how economic topics like sanctions and trade disruptions resonate differently among country groups (C2, C5, C9, and C17) based on regional interests and alliances, with Western nations (C2) often intensifying discussions on sanctions, while other regions engage more with trade or neutral relations (Graph (a) Figure 12). Politically, the BAR platform captures the dynamic spread of topics like diplomacy, election interference, and military alliances, which tend to peak in alignment with geopolitical events, reflecting the divided international stance on the conflict (Graph (b) Figure 12). The geographical barrier, represented in Figure (Graph (c) Figure 12), surfaces pressing humanitarian issues such as refugee flows and supply chain disruptions, illustrating the physical impact of the conflict on bordering regions and beyond. This approach aligns with previous studies using platforms like GDELT, which enables real-time analysis of global news events (Dominguez-Catena et al., 2024; Iglesias et al., 2016; Leetaru and Schrodt, 2013), and Event Registry, known for tracking cross-border news sentiment and media bias (Leban et al., 2014). Together, these insights suggest that the BAR-analytical platform provides a powerful framework for capturing the complex, multi-faceted impacts of international conflicts, offering a granular view that supports informed decision-making across economic, political, and humanitarian dimensions.

- **RO1. Analyze the differences in propagation, trends, sentiment, and topics across various barriers for both conflicts.** When comparing both conflicts, analysis shows that trends follow similar patterns of reporting. Interestingly, sentiment of news is way more negative for the Israel-Palestine war for all three barriers, compared to rather more positive sentiment regarding Ukraine-Russian war. Both wars share general trends such as economical trends and diplomacy, but differ in how human rights violations are emphasized in the Israel-Palestine war, but election interference in the Ukraine-Russian war.

- **RO2. Determine the similarities and differences among events of the same nature, such as wars.** To summarize, both wars share similarities in reporting (trends, propagation). When we look in the details of how the events are reported, different topics emerge, especially for political barriers, with addition of strike differences in sentiment reporting: more negative for Israeli-Palestine, compared to more positive for Russia-Ukraine war.

## 8. Research Limitations

While this study offers valuable insights into the differences in news reporting across geographical, economic, political, and cultural boundaries, there are several limitations that should be acknowledged:

The analysis in this paper is based on news articles from selected countries and media outlets. Although the study aims to cover a broad range of perspectives, it is limited by the availability and selection of news sources. Some smaller or less prominent news outlets may not be fully represented, which could impact the comprehensiveness of the analysis.

The research covers the Israeli-Palestinian and Russia-Ukraine conflicts during specific periods. The dynamic nature of ongoing conflicts means that the trends, sentiments, and topics identified may evolve over time. A broader or more extended time frame might yield different patterns or emphasize other aspects of these conflicts.

Although efforts were made to include a diverse range of political and economic perspectives, media outlets inherently carry biases that can influence their reporting. These biases could be more pronounced in some countries or regions, limiting the objectivity of the findings and potentially skewing the overall results.

While sentiment analysis was used to provide insights into the emotional tone of the news articles, it has not been explicitly validated. The sentiment analysis model, although effective, may not have captured the full complexity of sentiment in the articles, and the lack of direct validation introduces a limitation. Without further validation or comparison with human-coded sentiment or other metrics, the robustness of the sentiment analysis results could be compromised, affecting the reliability of sentiment-driven insights.

The clustering analysis was conducted based on hierarchical clustering using topic modeling, but additional clustering validation techniques, such as silhouette scores or other internal validation metrics, were not performed. While topic coherence and topic diversity scores were utilized to assess the relevance and coverage of the identified topics, assessing clustering quality with additional metrics such as silhouette scores or external validation methods could provide further confidence in the accuracy and meaningfulness of the clusters.

## 9. Conclusions and Future work

In this paper, we introduced BAR-Analytics, a comprehensive analytical platform that enables systematic examination of news dissemination across various barriers - economic, political, and geographical. Through its multi-dimensional approach, BAR-Analytics reveals distinct differences in how events, particularly the Russian-Ukrainian and Israeli-Palestinian conflicts, are reported and interpreted across regions, economies, and political landscapes. Our case study demonstrated that the topics of these conflicts are portrayed uniquely based on regionally economic contexts, political orientations, and geographical proximity, with Western media often emphasizing geopolitical stakes and humanitarian crises, while non-Western media may focus on regional security and economic implications. The results underline the significance of barriers in shaping public perception and highlight the critical need for an integrated perspective on global news dynamics, as support provided by the proposed BAR-Analytics platform. The proposed platform also adds a valuable dimension to media studies, supporting more granular, data-driven understanding of international news coverage across cultural and regional divides.

Future directions for BAR-Analytics include expanding the platform's analytical capabilities to incorporate real-time updates and broader media channels, such as social media and blogs, which play an increasing role in news dissemination. Through these enhancements, BAR-Analytics can evolve into an indispensable resource for journalists, researchers, and policymakers aiming to monitor and interpret the complex, multi-dimensional landscape of global news.

## 9.1. Acknowledgment

The research presented in this paper was funded by the Slovenian Research Agency through project J2-1736 Causalify, the Ministry of Digital Transformation and Slovenian Research and Innovation Agency under CRP V2-2272 and V5-2264, as well as the EU's Horizon Europe Framework under grant agreement number 101095095 and 101094905 (AI4Gov).

# A. Propagation Analysis

## A.1. Israeli-Palestine war

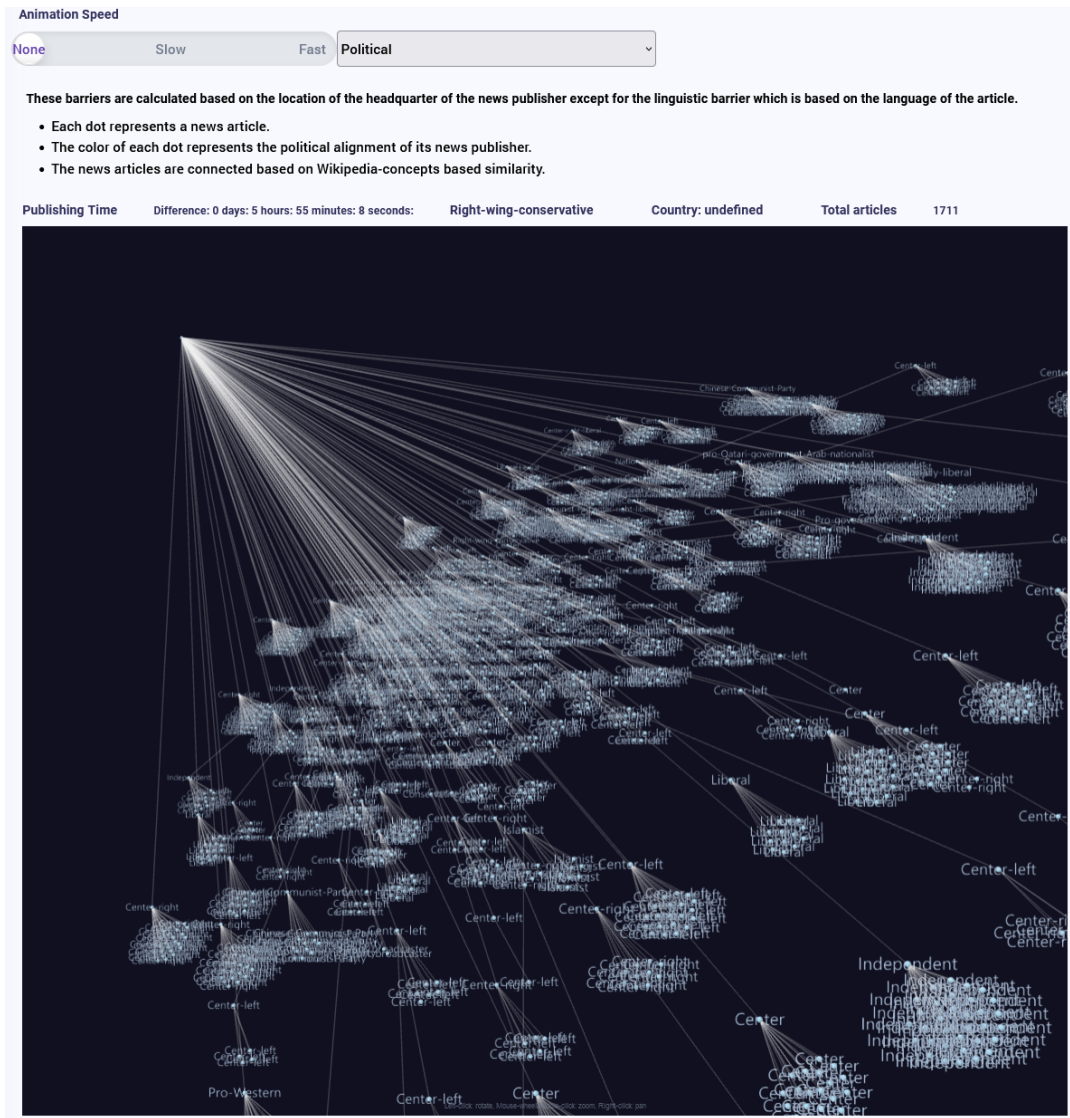

**Figure 13:** The network graph illustrates the propagation network of news articles published in the month of November 2023 related to Israeli-Palestine war and it represents propagation across political barrier.

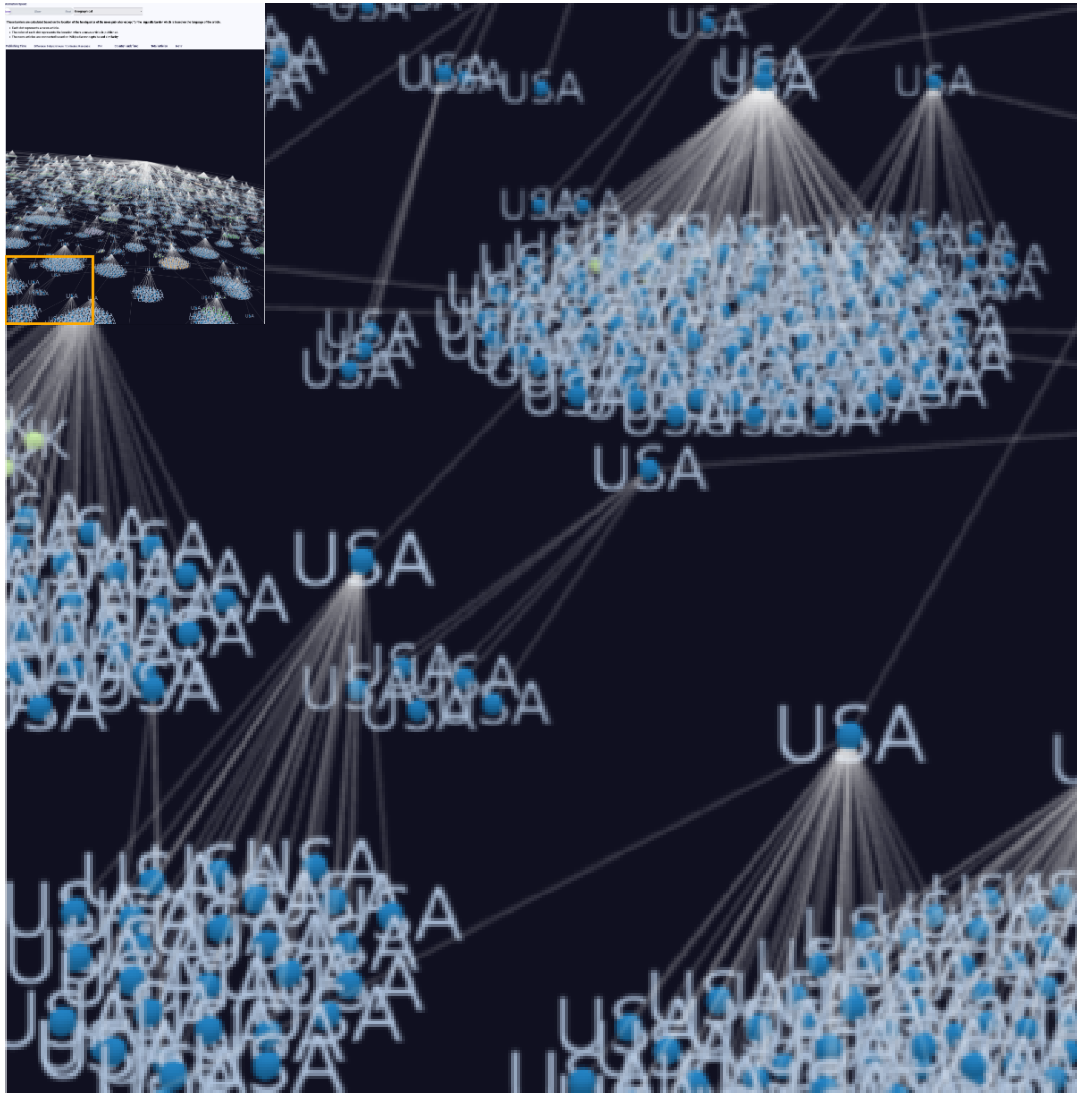

**Figure 14:** The network graph illustrates the propagation network of news articles published in the month of November 2023 related to Israeli-Palestine war and it represents propagation across geographical barrier. See close-up for detailed view of how a news from USA gets propagated mostly in USA.

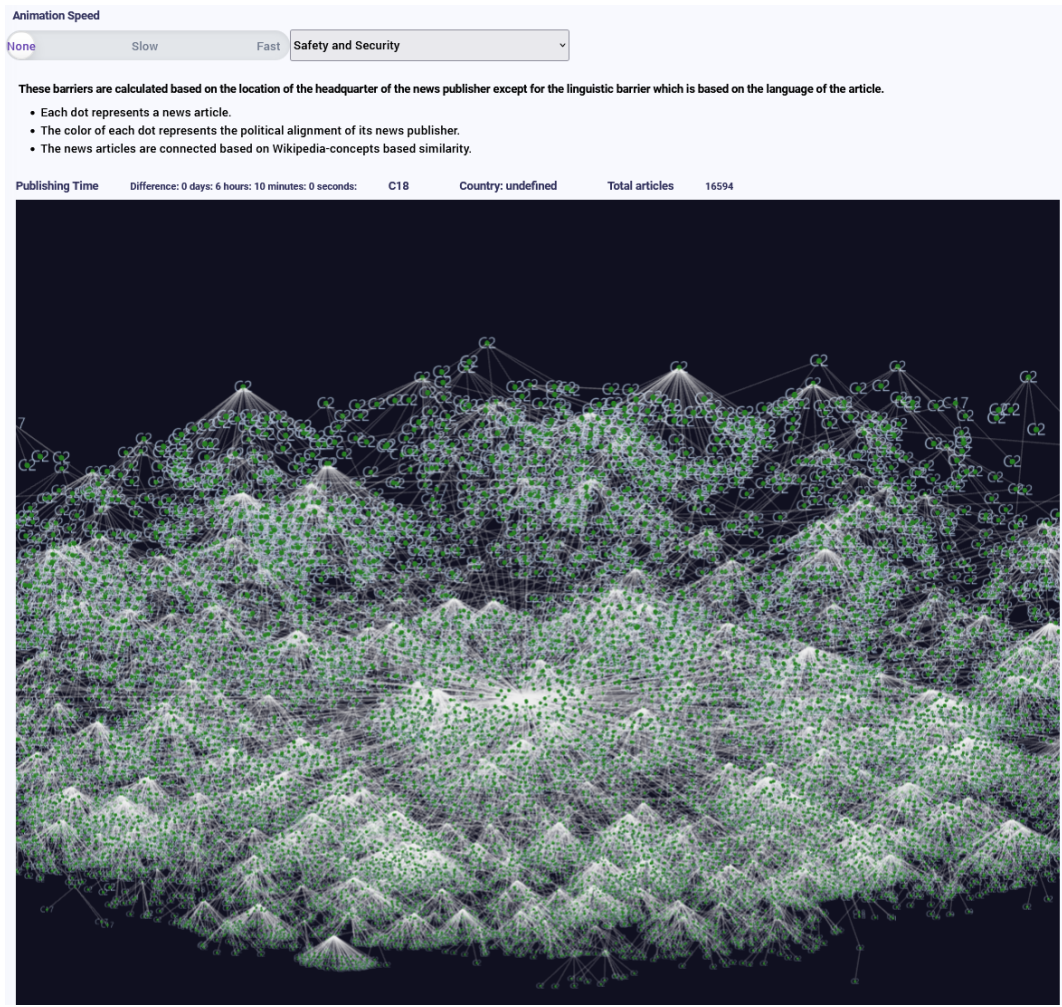

**Figure 15:** The network graph illustrates the propagation network of news articles published in the month of November 2023 related to Israeli-Palestine war and it represents propagation across economic barrier.

### A.2. Russia–Ukraine war

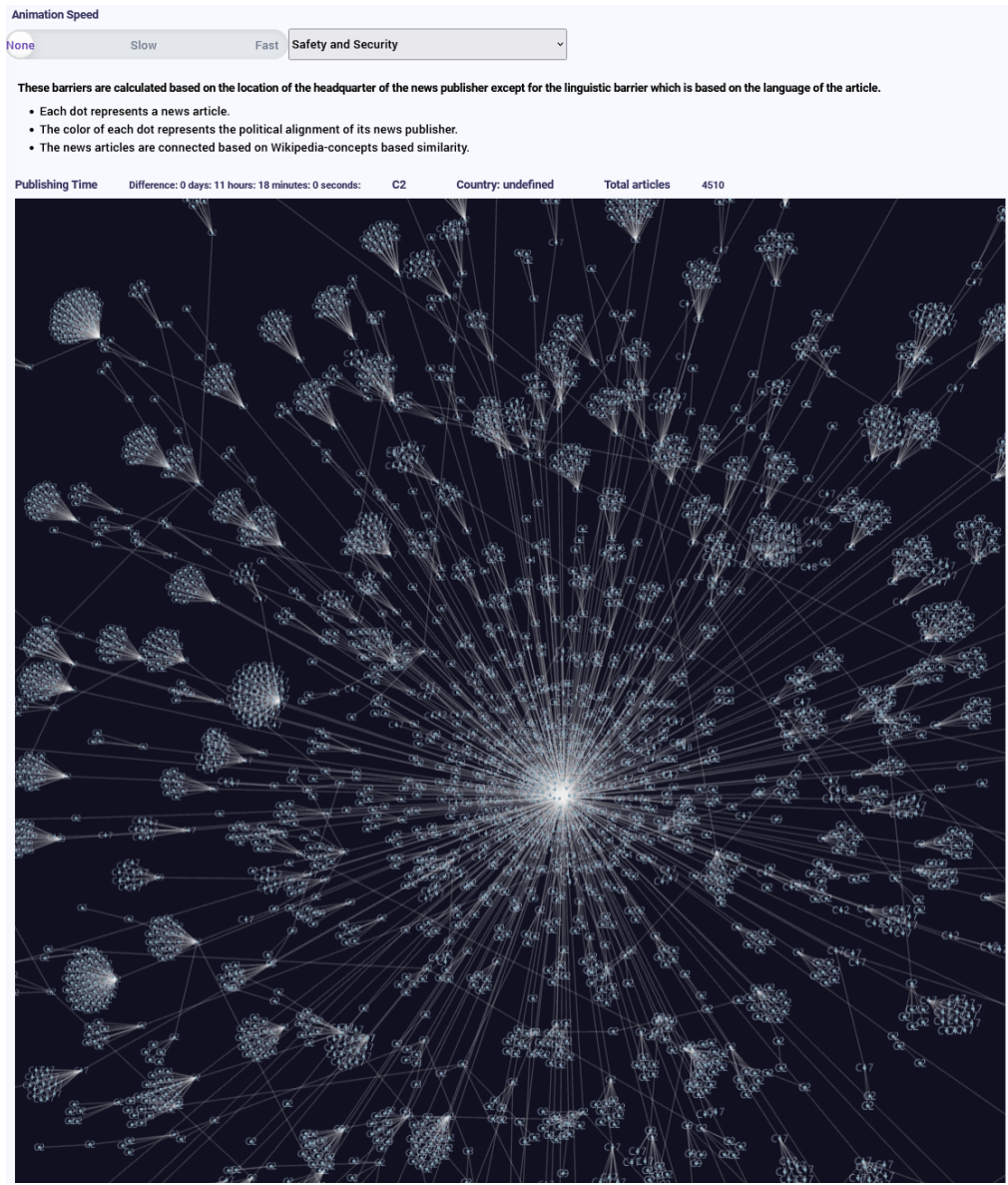

**Figure 16:** The network graph illustrates the propagation network of news articles published in the month of April 2023 related to Russian-Ukraine war and it represents propagation across economic barrier.

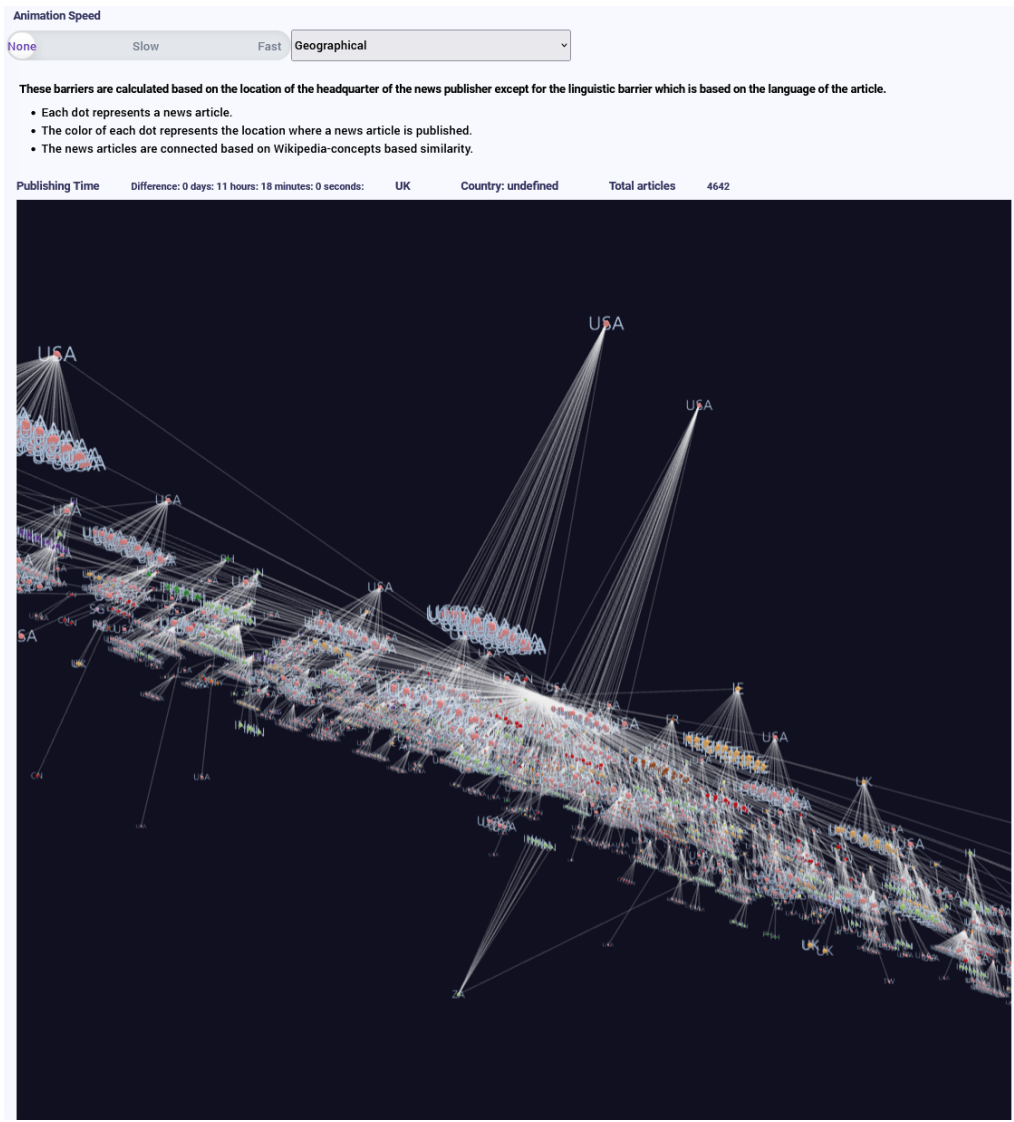

**Figure 17:** The network graph illustrates the propagation network of news articles published in the month of April 2023 related to Russian-Ukraine war and it represents propagation across geographical barrier.

## B. Trends Analysis

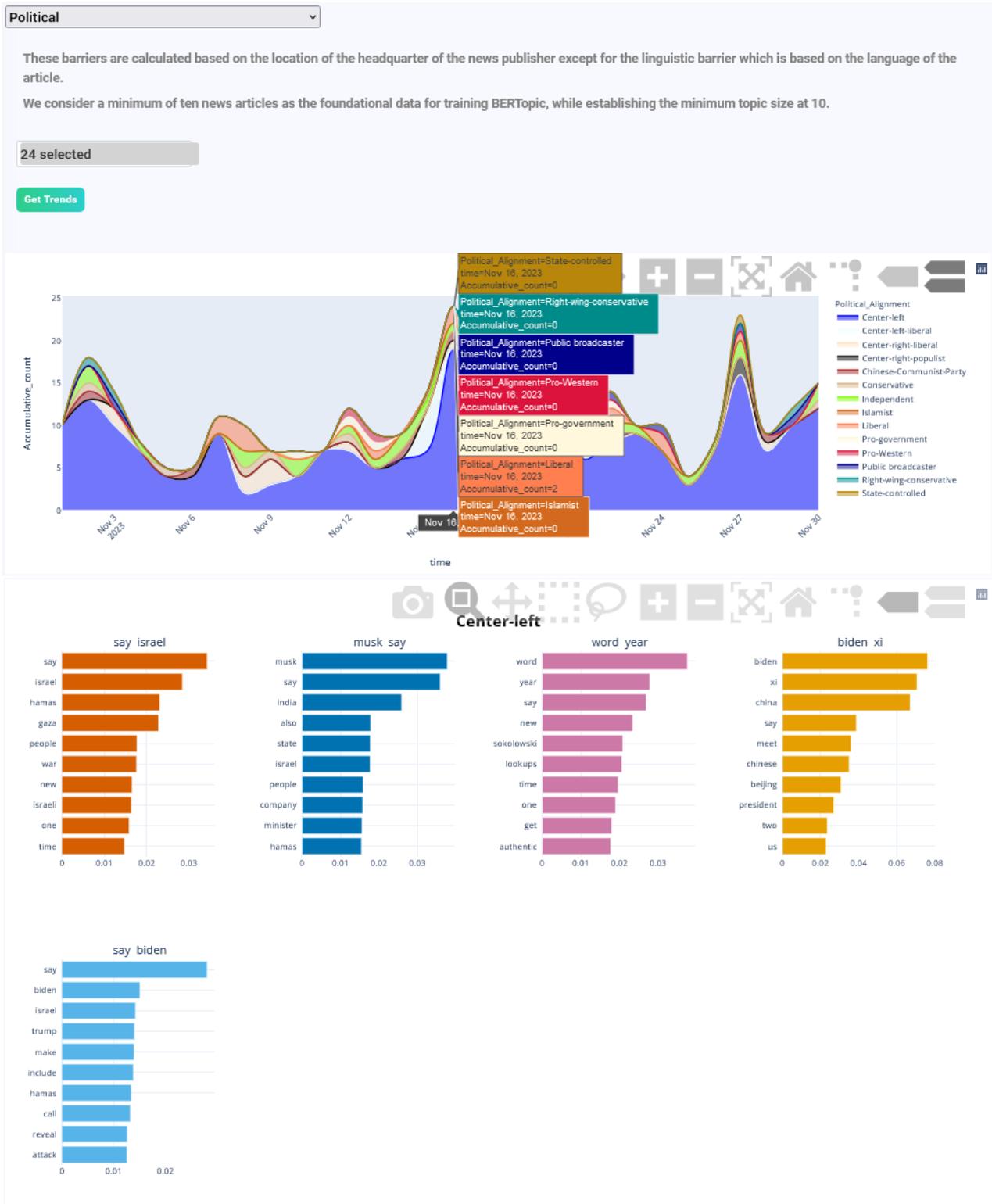

**Figure 18:** The line graphs illustrate the trends of news articles across political barrier published in the month of November 2023 related to Israeli-Palestine war.

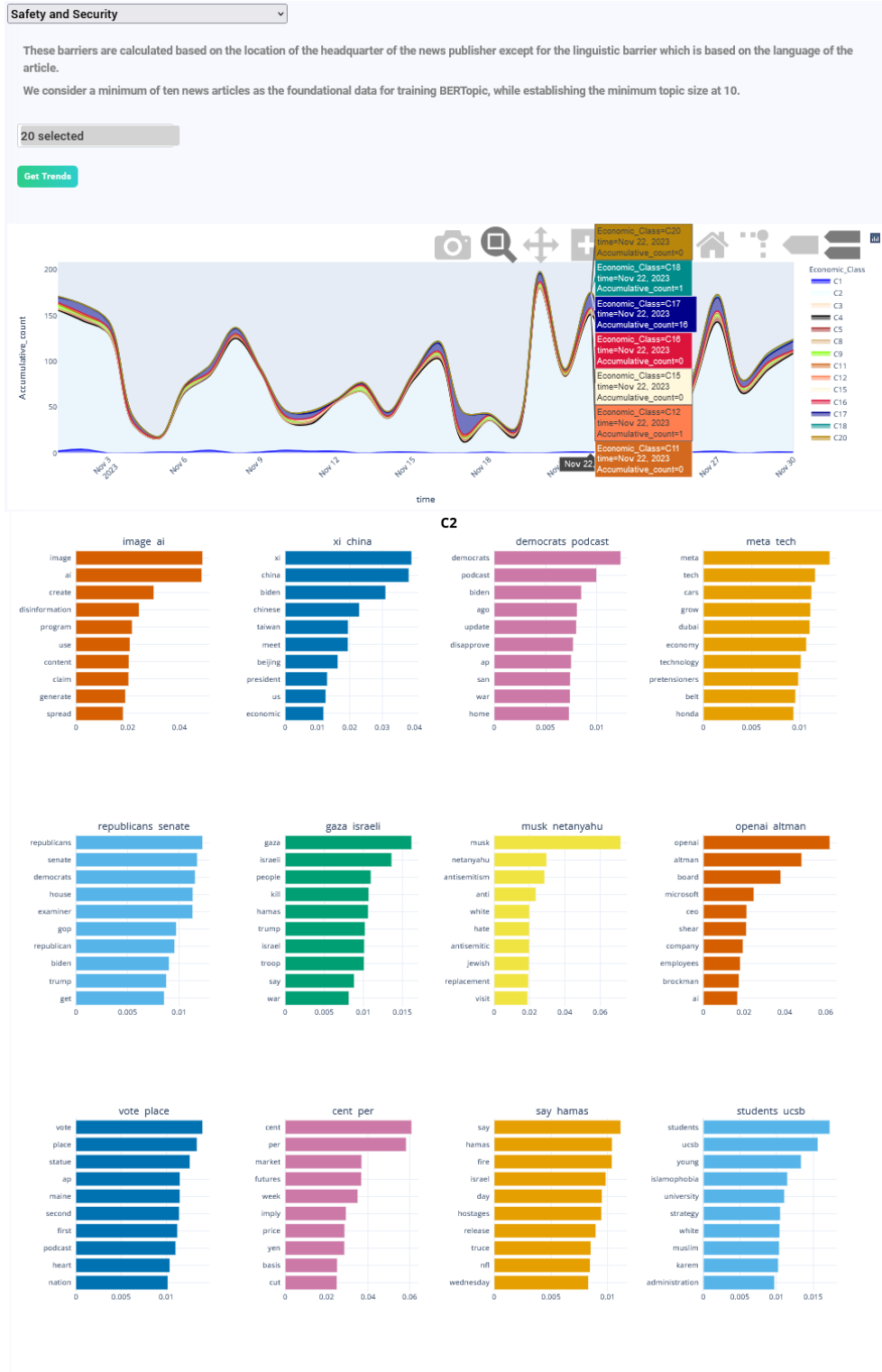

**Figure 19:** The line graphs illustrate the trends of news articles across economic barrier published in the month of November 2023 related to Israeli-Palestine war.

## C. Topics Analysis

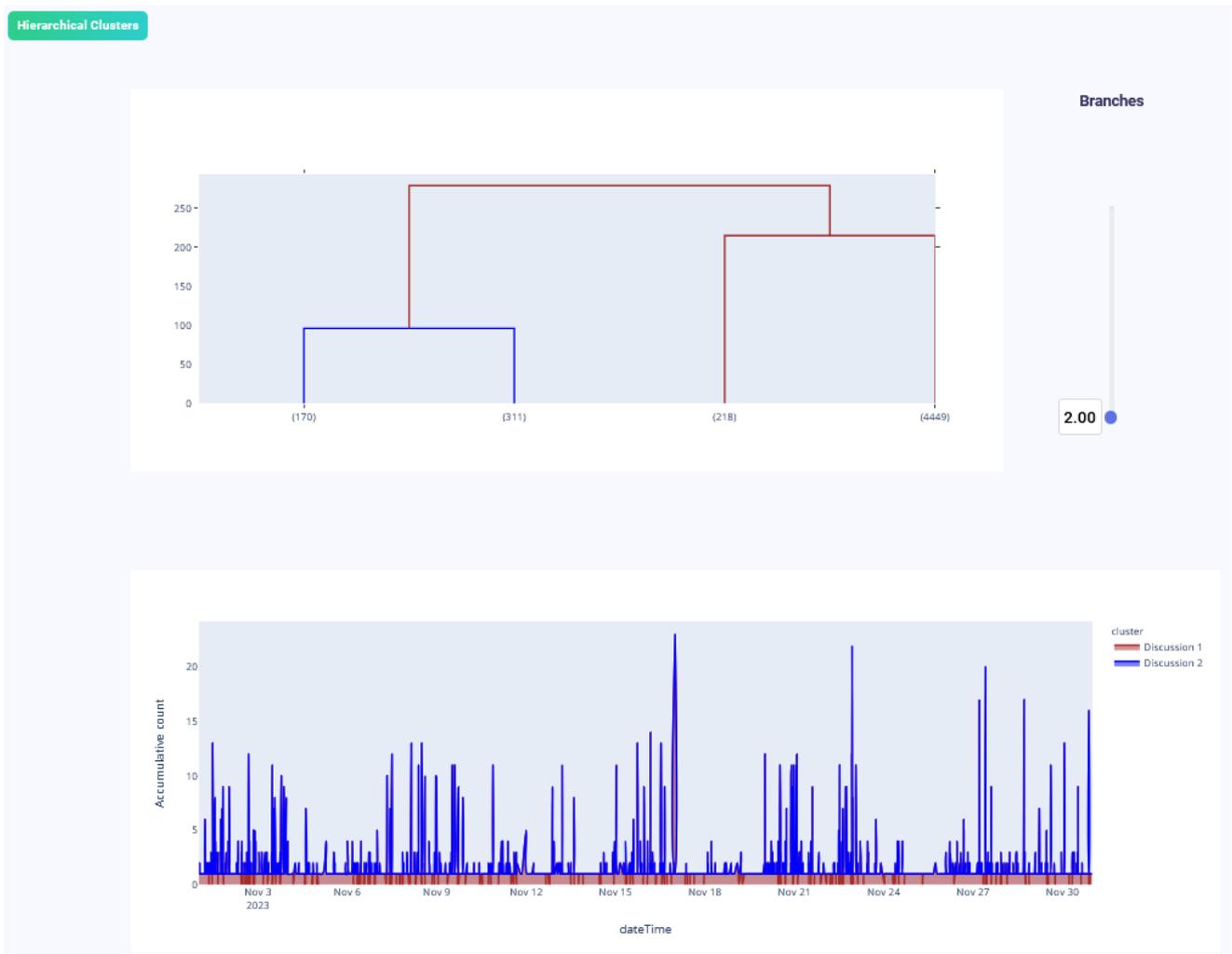

**Figure 20:** The bar charts show the frequent topics across economic barrier published in the month of November 2023 related to Israeli-Palestine war (1).

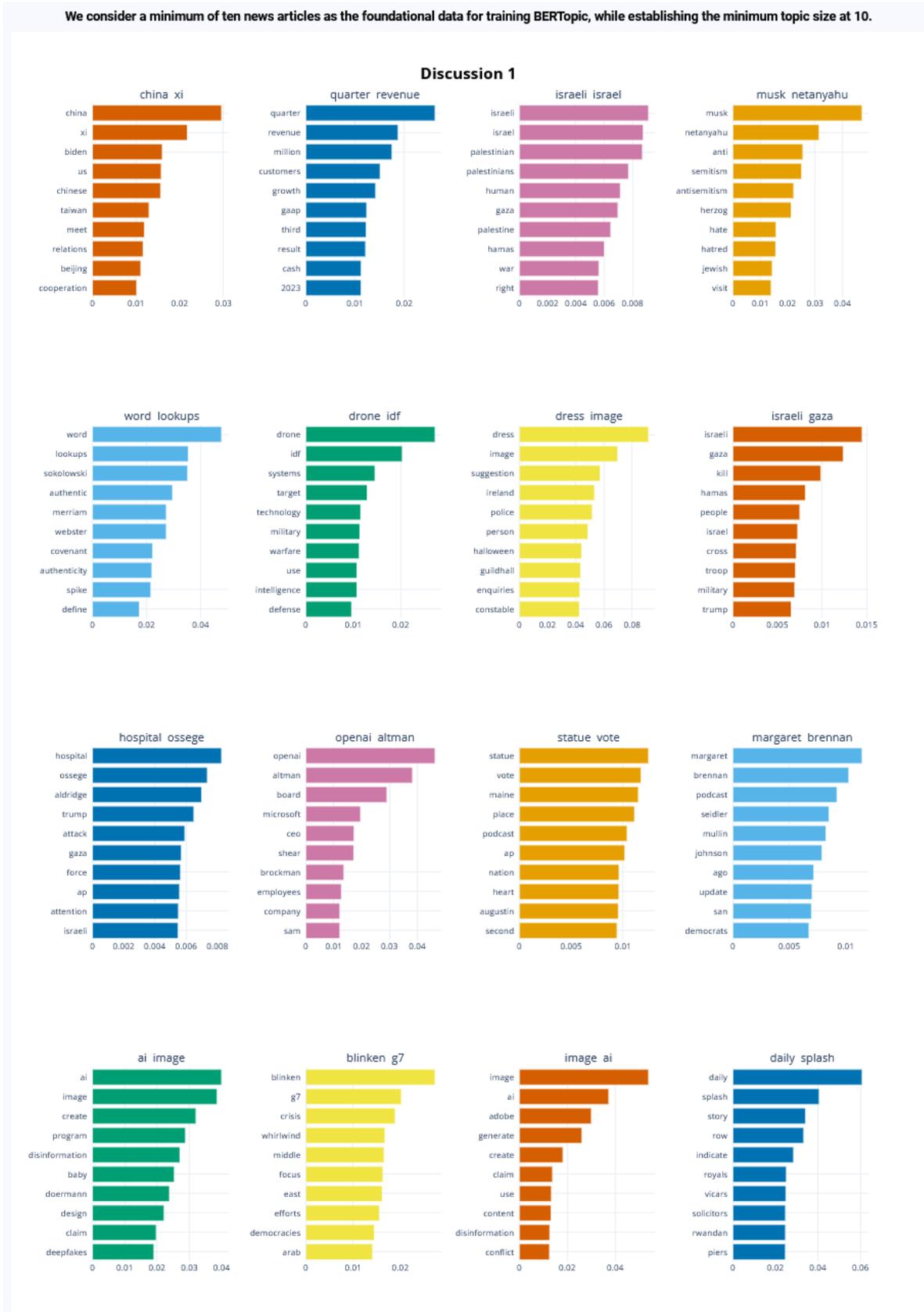

**Figure 21:** The bar charts show the frequent topics across economic barrier published in the month of November 2023 related to Israeli-Palestine war (2).

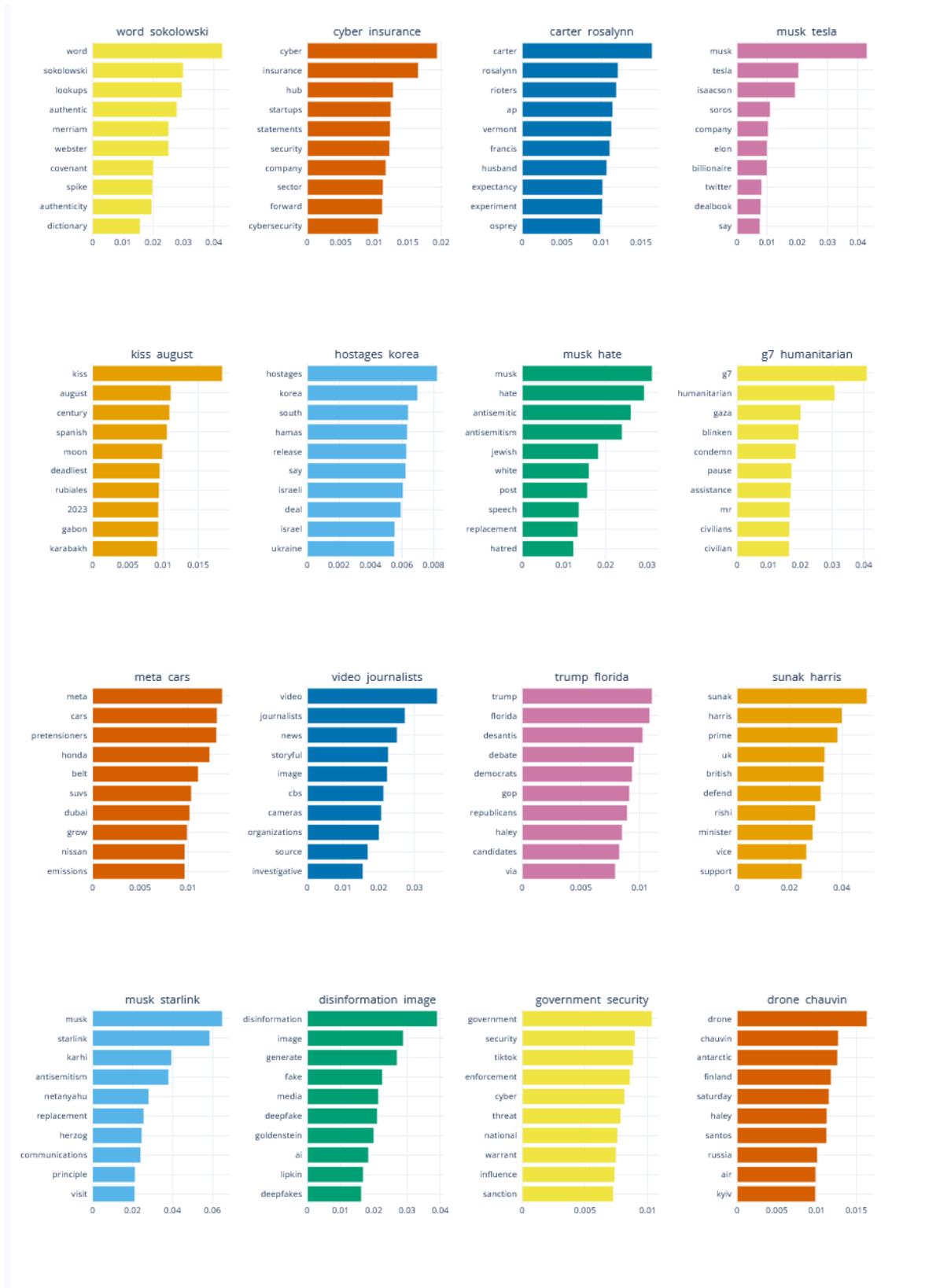

**Figure 22:** The bar charts show the frequent topics across economic barrier published in the month of November 2023 related to Israeli-Palestine war (3).

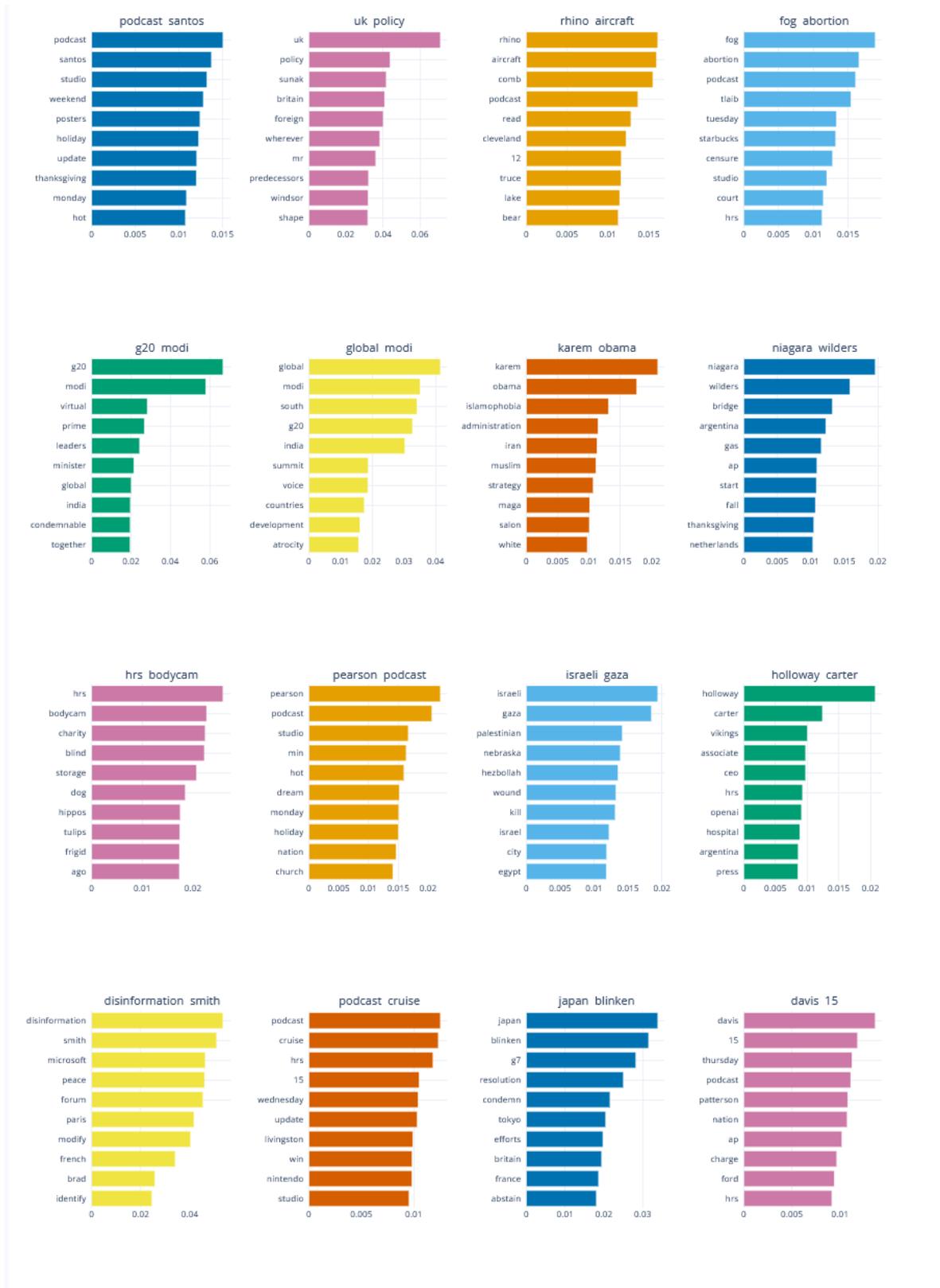

**Figure 23:** The bar charts show the frequent topics across economic barrier published in the month of November 2023 related to Israeli-Palestine war (4).

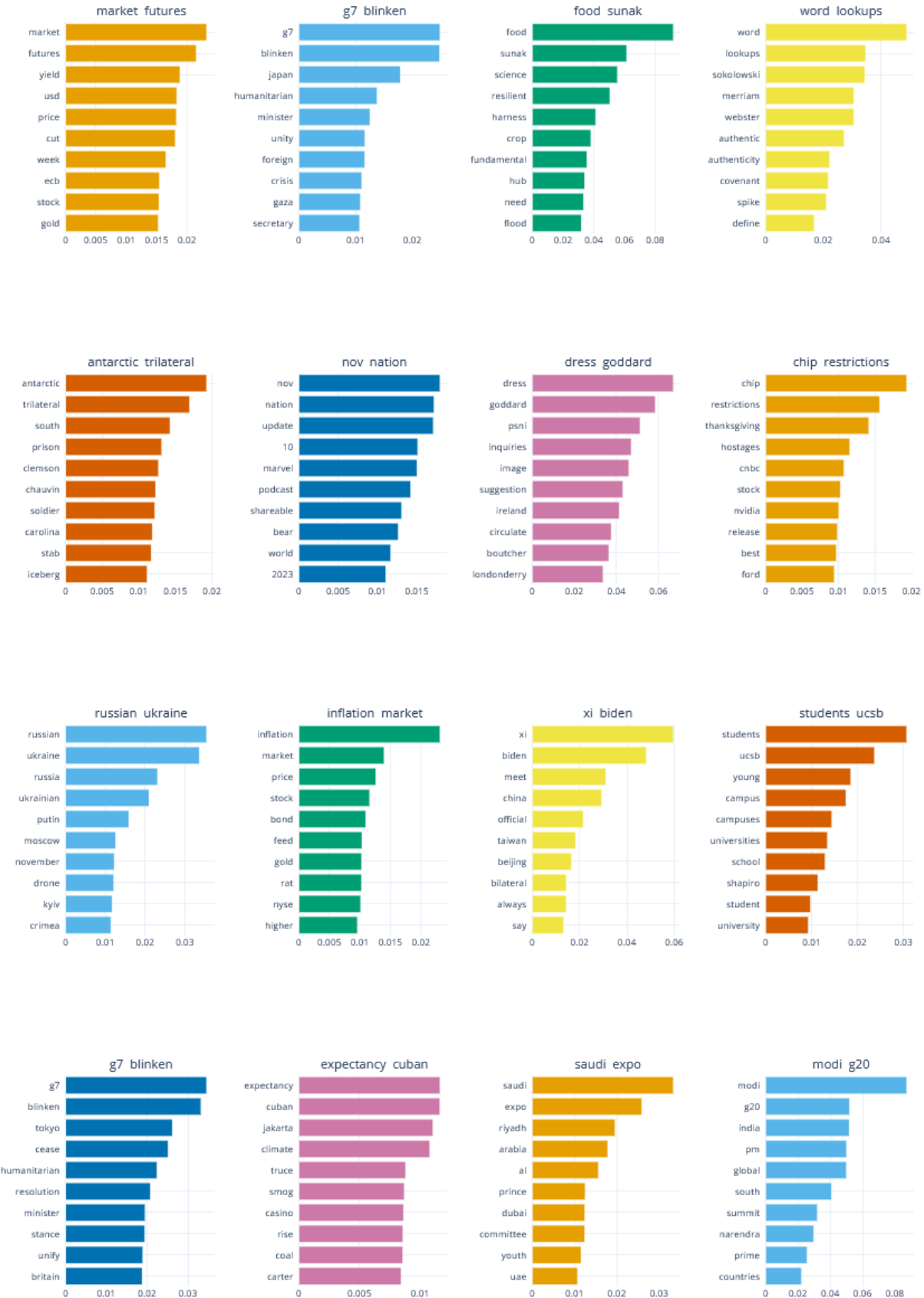

**Figure 24:** The bar charts show the frequent topics across economic barrier published in the month of November 2023 related to Israeli-Palestine war (5).

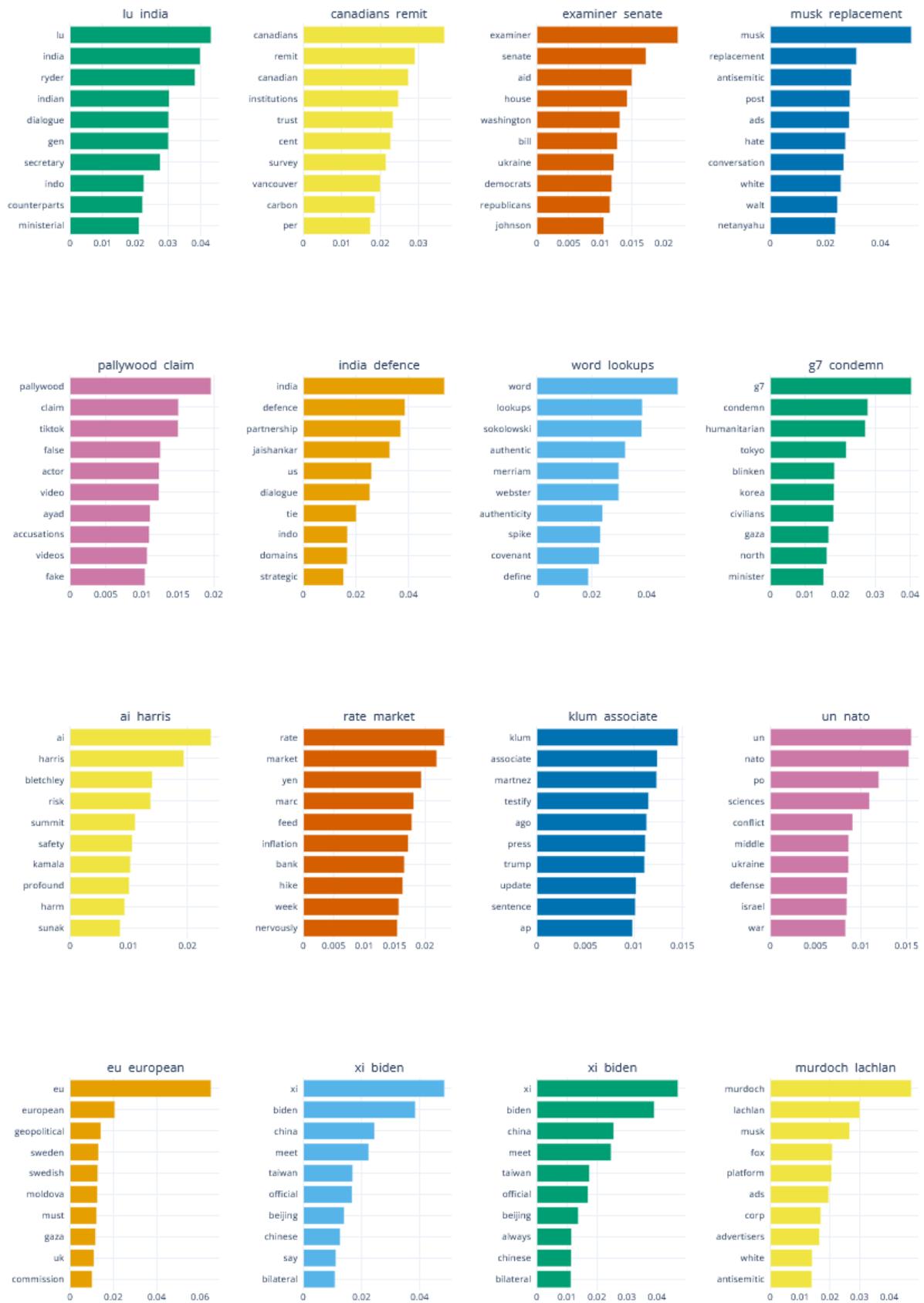

**Figure 25:** The bar charts show the frequent topics across economic barrier published in the month of November 2023 related to Israeli-Palestine war (6).

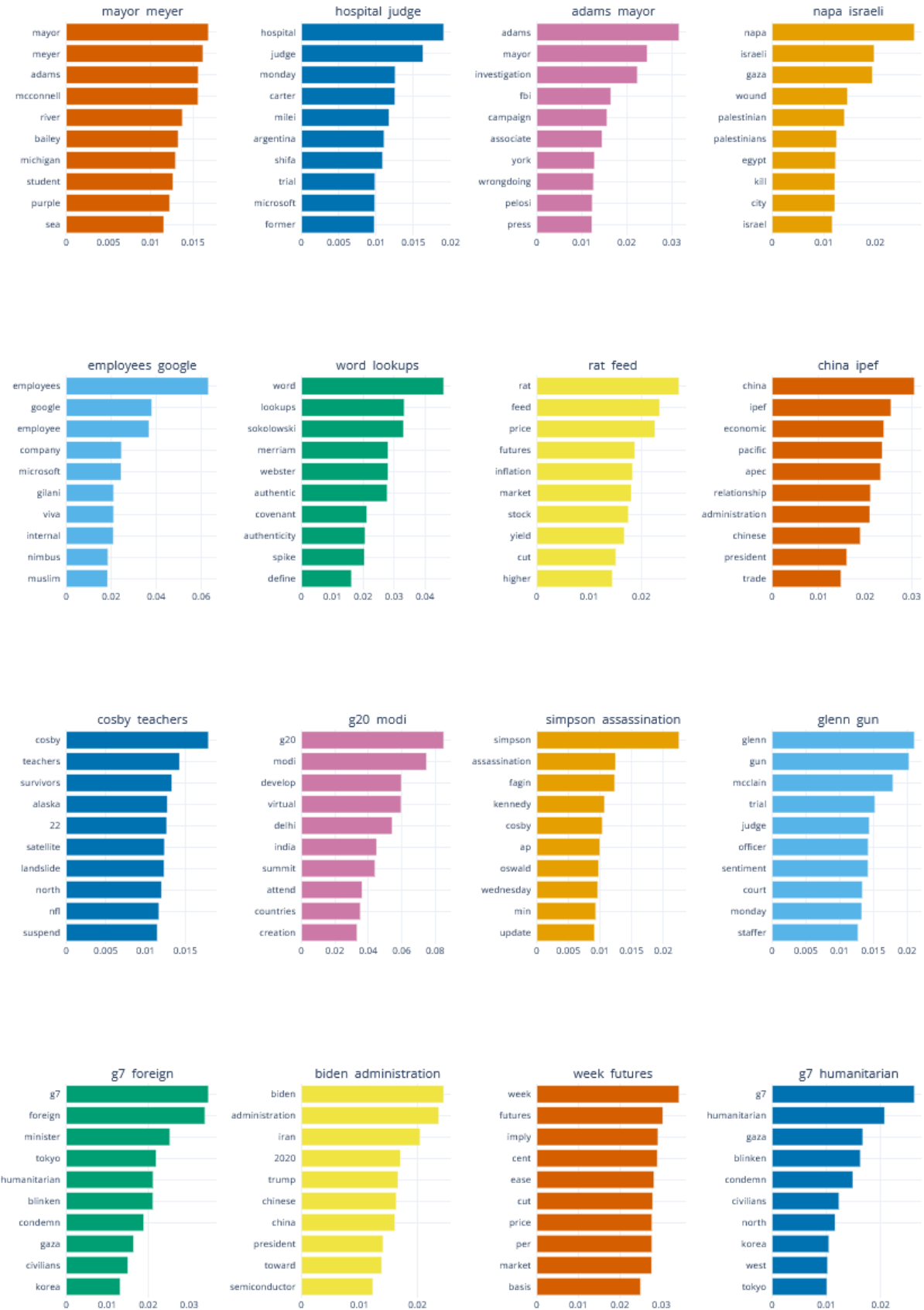

**Figure 26:** The bar charts show the frequent topics across economic barrier published in the month of November 2023 related to Israeli-Palestine war (7).

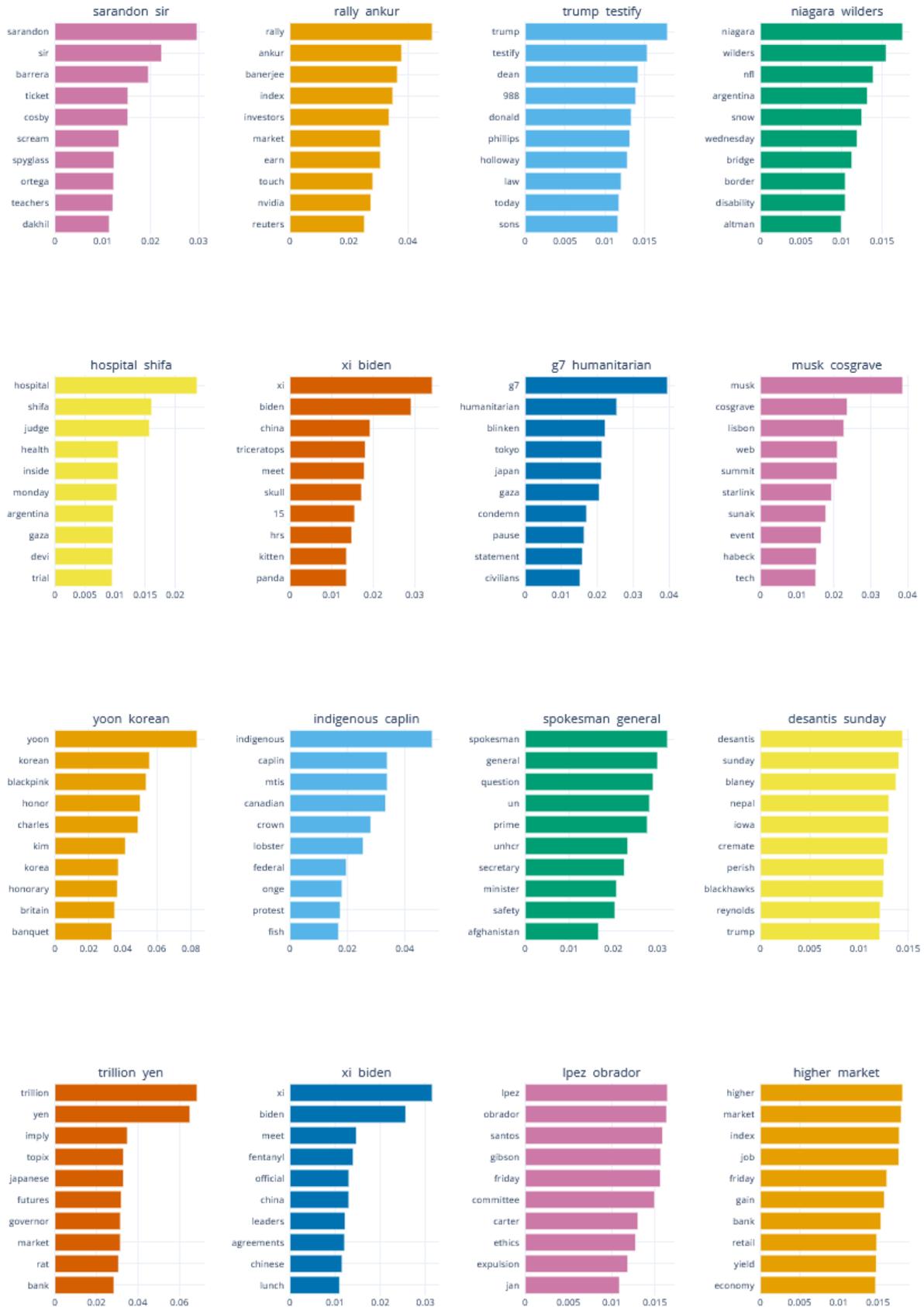

**Figure 27:** The bar charts show the frequent topics across economic barrier published in the month of November 2023 related to Israeli-Palestine war (8).

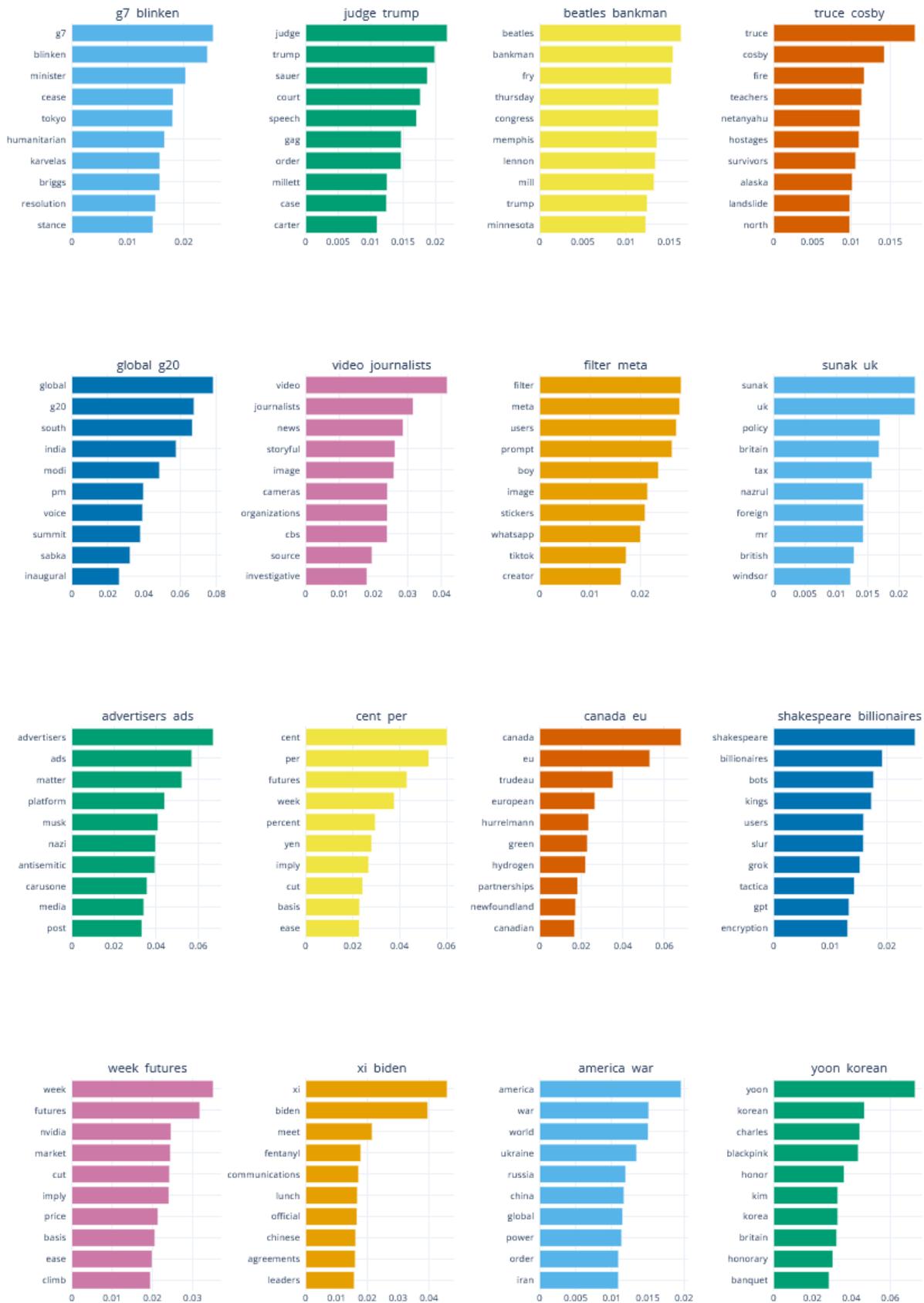

**Figure 28:** The bar charts show the frequent topics across economic barrier published in the month of November 2023 related to Israeli-Palestine war (9).

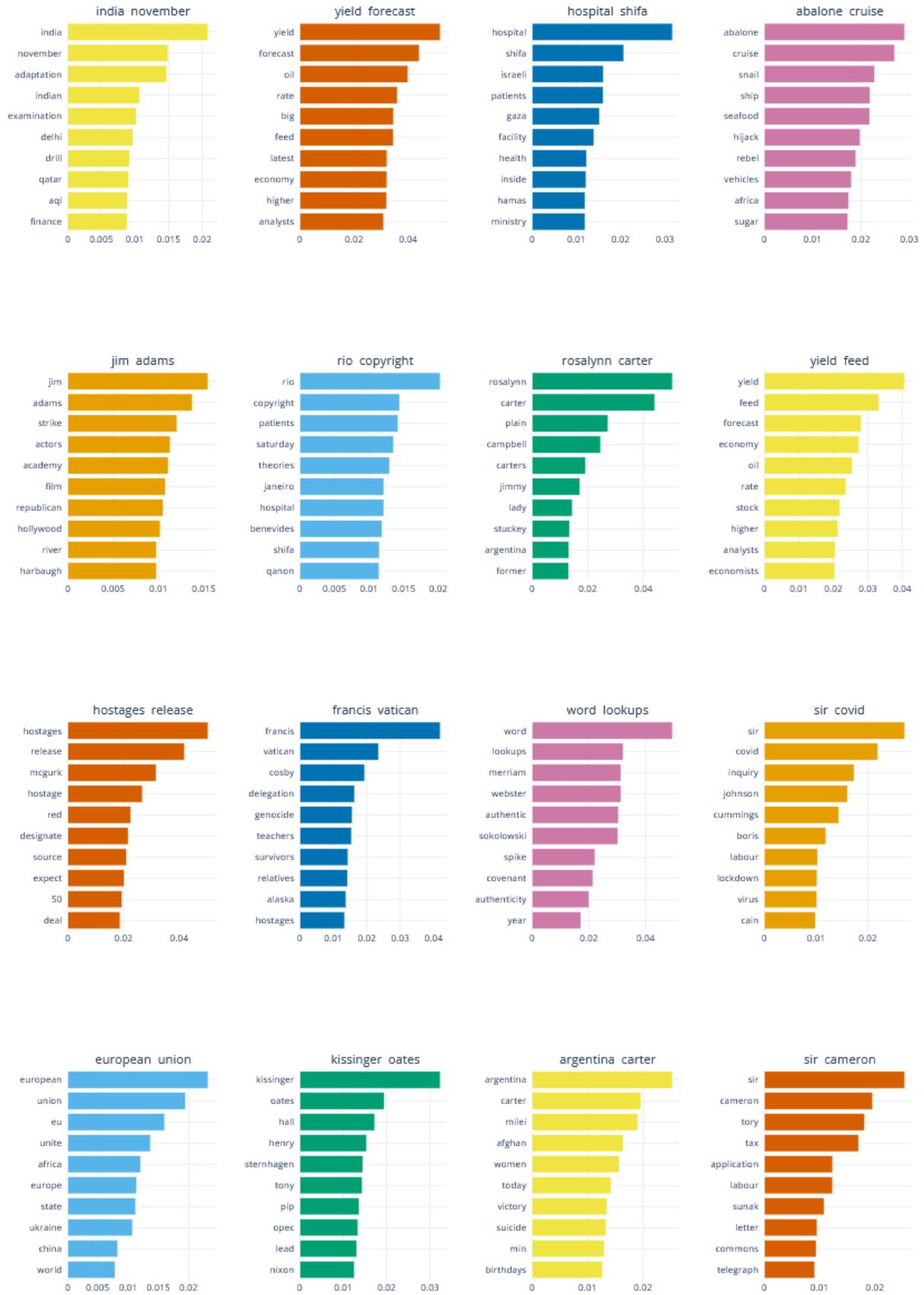

**Figure 29:** The bar charts show the frequent topics across economic barrier published in the month of November 2023 related to Israeli-Palestine war (10).

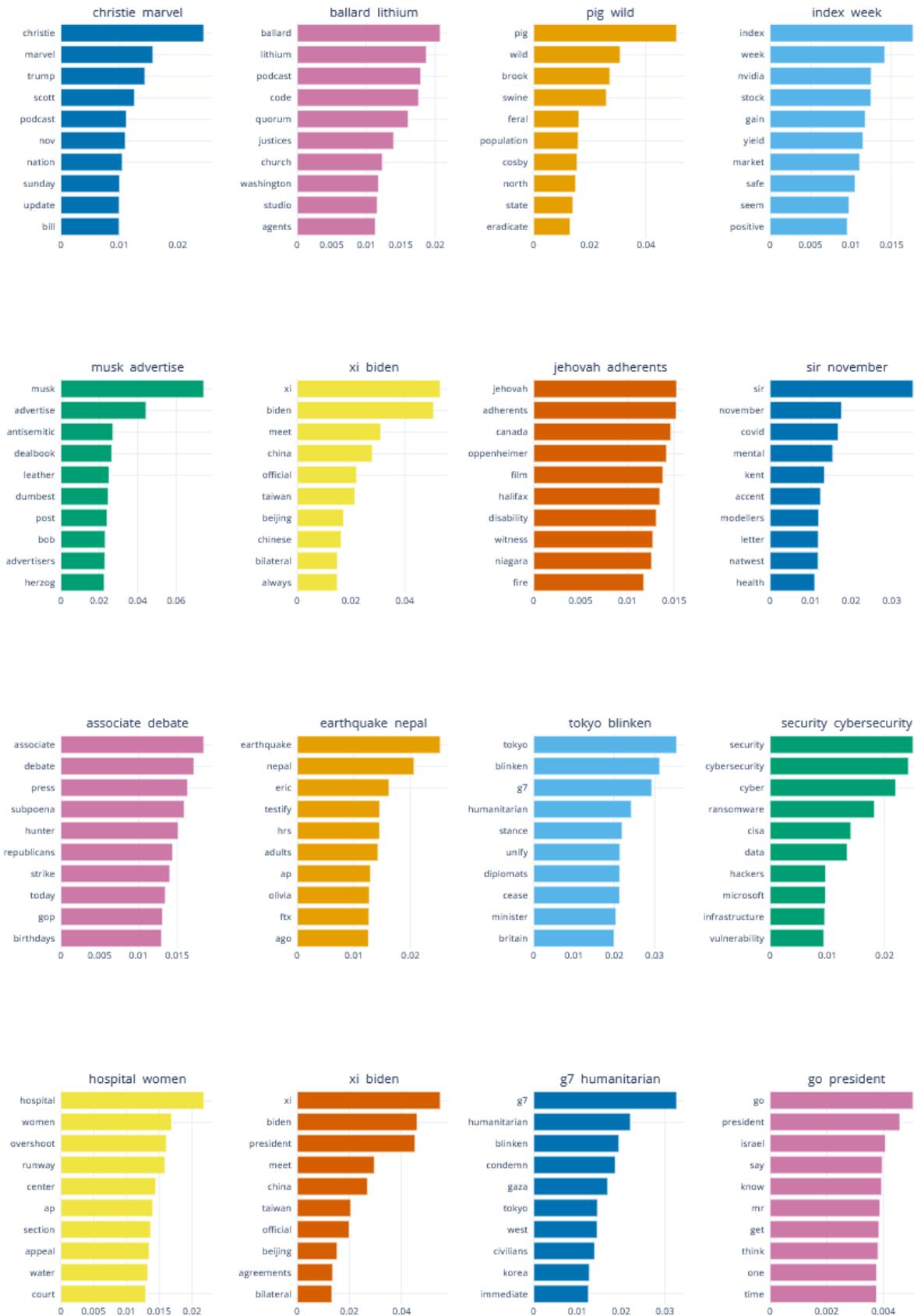

**Figure 30:** The bar charts show the frequent topics across economic barrier published in the month of November 2023 related to Israeli-Palestine war (11).

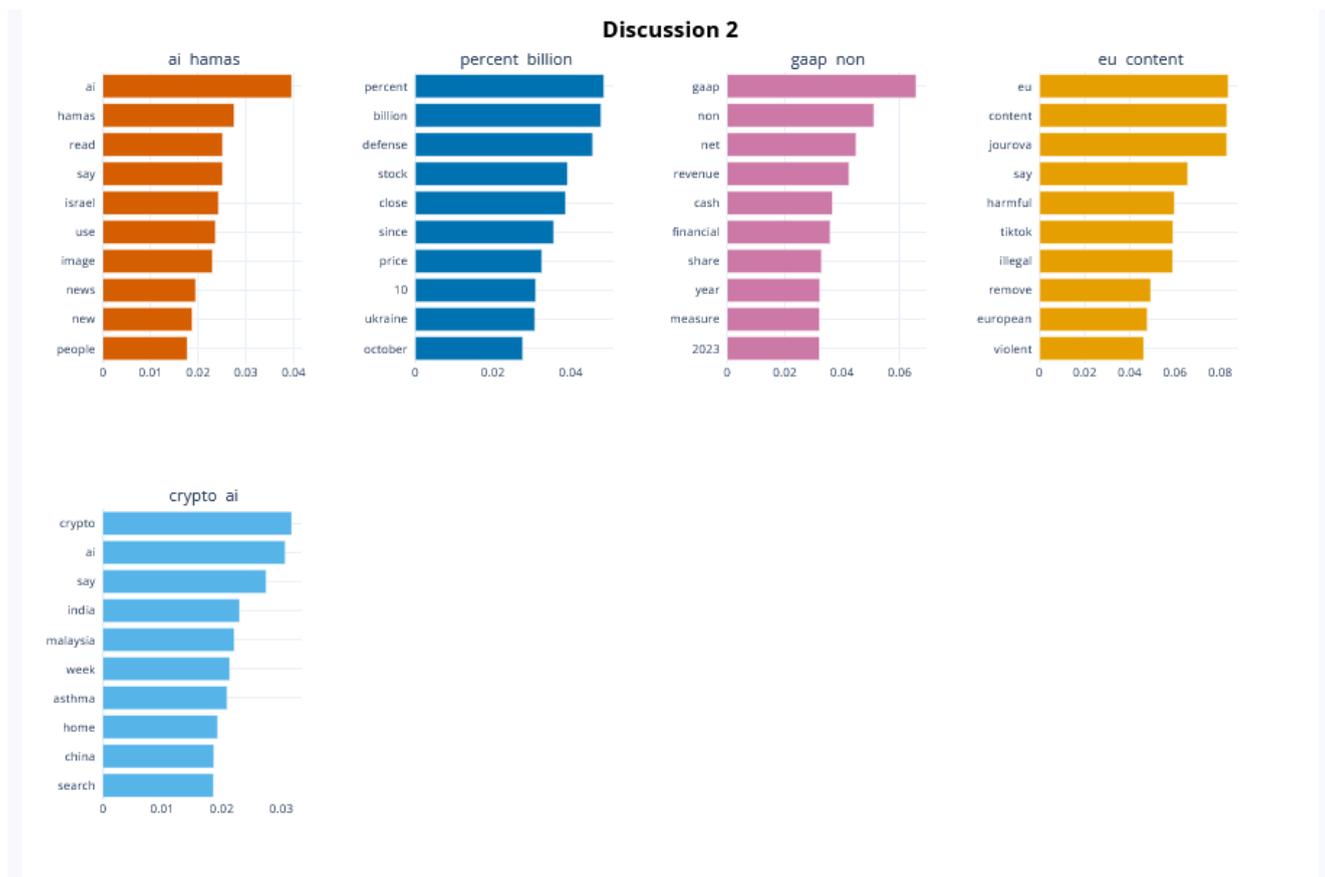

**Figure 31:** The bar charts show the frequent topics across economic barrier published in the month of November 2023 related to Israeli-Palestine war (12).

**Table 1**
**Clusters of countries with similar economies.**

| Classes | Countries | Classes | Countries | Classes | Countries |
|---|---|---|---|---|---|
| C1 | Israel | C8 | Afghanistan, Angola, Burundi, Cameroon, Central African Republic, Chad, Democratic Republic of the Congo, Eritrea, Ethiopia, Haiti, Mali, Mauritania, Republic of the Congo, Somalia, Sudan, Togo, Yemen, Zimbabwe | C15 | Jordan, Lebanon, Moldova, Morocco, Tunisia, Ukraine |
| C2 | Australia, Austria, Belgium, Canada, Denmark, Estonia, Finland, France, Germany, Hong Kong, Iceland, Ireland, Italy, Luxembourg, Malta, Netherlands, New Zealand, Norway, Portugal, Singapore, Slovenia, Spain, Sweden, Switzerland, Taiwan, United Kingdom, United States | C9 | Azerbaijan, Bahrain, China, Kazakhstan, Kuwait, Malaysia, Oman, Qatar, Saudi Arabia, Thailand, United Arab Emirates, Vietnam | C16 | Ghana, Namibia, Senegal, The Gambia |
| C3 | Czech Republic, Japan, Korea, South Korea | C11 | Cambodia, Equatorial Guinea, Guatemala, Honduras, Laos, Myanmar, Nicaragua | C17 | Botswana, El Salvador, India, Kenya, Rwanda, South Africa |
| C4 | Armenia, Bulgaria, Chile,Costa Rica, Croatia, Cyprus, Georgia, Greece, Hungary, Latvia, Lithuania, Mauritius, Montenegro, Panama, Poland, Romania, Serbia, Seychelles, Slovakia, Uruguay | C12 | Belarus, Russia, Turkey | C18 | Colombia, Dominican Republic, Indonesia, Mexico, Philippines, Sri Lanka |
| C5 | Bangladesh, Djibouti, Egypt, Iran, Iraq, Libya, Nepal, Nigeria, Pakistan, Syria, Uganda | C13 | Algeria | C19 | Albania, Argentina, Brazil, Peru, Republic of Macedonia |
| C6 | Benin, Burkina Faso, Costa Rica, Guinea, Guinea-Bissau, Liberia, Madagascar, Malawi, Mozambique, Niger, Papua New Guinea, Sierra Leone,  Tanzania, Zambia | C14 | Jamaica | C20 | Bolivia, Bosnia and Herzegovina, Ecuador, Guyana, Kyrgyzstan, Mongolia, Paraguay, Suriname, Trinidad and Tobago |
| C7 | Gabon, Venezuela | | | | |